\documentclass[10pt,twocolumn,letterpaper]{article}

\usepackage[pagenumbers]{cvpr} 

\usepackage{graphicx}
\usepackage{amsmath}
\usepackage{amssymb}
\usepackage{color}
\usepackage[table]{xcolor}
\usepackage{multirow,array,varwidth}
\usepackage{booktabs}
\usepackage{makecell}
\usepackage{bbm}
\usepackage{dsfont}
\usepackage{xspace}
\usepackage{amsthm}
\usepackage{float}
\usepackage[normalem]{ulem}

\usepackage{enumitem}

\newcommand{\train}{{\small\texttt{train}\xspace}}
\newcommand{\val}{{\small\texttt{val}\xspace}}
\newcommand{\test}{{\small\texttt{test}\xspace}}

\newcommand{\knn}{$k$-NN\xspace}
\newcommand{\base}{\textsc{Base}}
\newcommand{\baseprime}{\textsc{Base}$^{\prime}$}

\newcommand{\jointinf}{\textsc{Joint-inf}}
\newcommand{\joint}{\textsc{Joint}}

\renewcommand{\paragraph}[1]{\vspace{0.12cm}\noindent\textbf{#1}\xspace\xspace\xspace}
\newcommand{\cvpara}[1]{\vspace{-0.2cm}\paragraph{#1}}
\newcommand{\suppparagraph}[1]{\vspace{0.25cm}\noindent\textbf{#1}\xspace\xspace\xspace}

\let\vec\mathbf 
\newcommand{\R}[0]{\mathds{R}}
\newcommand{\eqcomma}[0]{\;\;,}
\newcommand{\eqdot}[0]{\;\;.}

\newcommand{\vit}[0]{ViT}
\newcommand{\rn}[0]{ResNet}
\newcommand{\swin}[0]{Swin}

\newcommand{\suplong}[0]{Supervised}
\newcommand{\supshort}[0]{Sup}
\newcommand{\mocotwo}[0]{MoCo v2}
\newcommand{\moco}[0]{MoCo v3}
\newcommand{\swav}[0]{SwAV}
\newcommand{\dino}[0]{DINO}
\newcommand{\btwins}[0]{BarlowTwins}
\newcommand{\moby}[0]{MoBY}
\newcommand{\simclr}[0]{SimCLR}

\newcommand{\inatlong}[0]{iNaturalist2021}
\newcommand{\inat}[0]{iNat2021}

\newcommand{\imagenet}[0]{ImageNet}
\newcommand{\longcub}[0]{Caltech-UCSD Birds-200-2011}
\newcommand{\cub}[0]{CUB-200-2011}
\newcommand{\flowers}[0]{Oxford Flowers}
\newcommand{\cars}[0]{Stanford Cars}
\newcommand{\dogs}[0]{Stanford Dogs}
\newcommand{\newt}[0]{NeWT}

\definecolor{green_im}{rgb}{0.1, 0.55, 0.3}
\newcommand{\meanstd}[2] { #1 \textsubscript{$\pm$  #2} }

\newcommand{\Rise}[1]{\textcolor{green_im}{\tiny{(\bf $\uparrow$#1})}\xspace}

\definecolor{dark_blue}{rgb}{10, 120, 180}

\definecolor{citecolor}{RGB}{0, 113, 188}

\usepackage[pagebackref,breaklinks,colorlinks,citecolor=citecolor]{hyperref}

\usepackage[capitalize]{cleveref}
\crefname{section}{Sec.}{Secs.}
\Crefname{section}{Section}{Sections}
\Crefname{table}{Table}{Tables}
\crefname{table}{Tab.}{Tabs.}

\def\confName{CVPR}
\def\confYear{2022}

\begin{document}

\title{Rethinking Nearest Neighbors for Visual Classification}

\author{
  Menglin Jia~$^{1,2}$\hspace{10pt}
  Bor-Chun Chen$^{2}$\hspace{10pt}
  Zuxuan Wu~$^{3}$\hspace{10pt}
  Claire Cardie$^{1}$\hspace{10pt}
  Serge Belongie$^{4}$\hspace{10pt}
  Ser-Nam Lim$^{2}$ \\
$^{1}$Cornell University
\qquad $^{2}$Meta AI
\qquad $^{3}$Fudan University
\qquad $^{4}$University of Copenhagen
}

\maketitle

\begin{abstract}

Neural network classifiers have become the de-facto choice for current ``pre-train then fine-tune''
paradigms of visual classification. 
In this paper, we investigate $k$-Nearest-Neighbor (\knn{}) classifiers, a classical model-free learning method from the pre-deep learning era, as an augmentation to modern neural network based approaches.
As a lazy learning method, \knn{} simply aggregates the distance between the test image and top-$k$ neighbors in a training set. 
We adopt \knn{} with pre-trained visual representations produced by either supervised or self-supervised methods in two steps: 
(1) Leverage \knn{} predicted probabilities as indications for easy \vs~hard examples during training.
(2) Linearly interpolate the \knn{} predicted distribution with that of the augmented classifier.
Via extensive experiments on a wide range of classification tasks, our study reveals the 
generality and flexibility 
of \knn{} integration with additional insights:
(1) \knn{} achieves competitive results, sometimes even outperforming a standard linear classifier.
(2) Incorporating \knn{} is especially beneficial for tasks where parametric classifiers perform poorly and / or in low-data regimes.
We hope these discoveries will encourage
people to rethink the role of pre-deep learning, classical methods in computer vision.\footnote{Our code is available at:
\href{https://github.com/KMnP/nn-revisit}{\texttt{github.com/KMnP/nn-revisit}}. }

\end{abstract}

\section{Introduction}\label{sec: intro}

Deep neural networks revolutionized computer vision, 
and have become a fundamental tool for a wide variety of applications.
With the advancement of representation learning, it is now common practice to 
leverage the features extracted from large-scaled labeled data (\eg, \imagenet{}~\cite{imagenet_cvpr09}),
or other pretext tasks 
(\eg, 
colorization~\cite{zhang2016colorful,zhang2017split}, 
rotations of inputs~\cite{gidaris2018rotnet}, 
modeling image similarity and dissimilarity among multiple views~\cite{he2020moco,chen2020mocov2,chen2020simclr,caron2020swav,chen2021mocov3, zbontar2021barlow, caron2021dino}, to name a few), and transfer them to target domains, such as
object detection~\cite{girshick2014rich,girshick2015fast,ren2015faster,he2017mask}, 
action recognition~\cite{simonyan2014two,carreira2017quo}, or various fine-grained recognition tasks~\cite{cui2018large,su2020does,VanHorn2021newt,ericsson2021well}.

\emph{Eager learners}~\cite{friedman2017elements}, like neural networks, require an intensive learning stage to estimate the model parameters. Thus the majority of computation occurs at training time.
A stark contrast to deep neural networks is the $k$-Nearest Neighbor (\knn{}) classifier: a classical, ``old-school'' learning method. 
As a \emph{lazy learning} approach, \knn{} requires no learning / training time. 
It simply memorizes training data and predicts labels based on the nearest training examples.
\knn{} also avoids over-fitting of parameters~\cite{boiman2008nbnn} and is robust to input perturbations~\cite{orhan2018simple}.
Because of the favorable properties of \knn{}, researchers (\eg~\cite{zhang2006svmknn,boiman2008nbnn,devlin2015exploring,wu2018improving,wu2018unsupervised,wang2019simpleshot,granot2021drop}) periodically revisit \knn{} each time new paradigms of eager learners emerge,
raising the questions: what role remains for this simple non-parametric learning approach? Is ``good old'' \knn{} still useful, or should it be relegated to historical background sections in pattern recognition textbooks?

In this paper, we investigate \emph{how to incorporate the complementary strengths of nearest neighbors and neural network methods} for current ``pre-train then fine-tune'' paradigms of visual classification. 
Our work builds upon recent findings that self-supervised Vision Transformer features enable excellent \knn{} classification~\cite{caron2021dino}. 
We further demonstrate that all features, regardless of the pre-training backbones or objectives, can benefit from \knn{}.
Using the embedded visual representations as input,
we propose to leverage the predictive results of a \knn{} classifier and augment conventional modern classifiers in two steps: 
(1) During training, we treat \knn{} results as an indication of easy vs.~hard examples in the training set. More specifically, we add a scaling factor to the standard cross-entropy criterion of conventional classifiers, automatically forcing the model to focus on the hard examples identified by \knn{}.
(2) During test time, we linearly interpolate probability distributions from the learned augmented model with the distributions produced by \knn{}.

Extensive experiments show that incorporating \knn{} improves standard parametric classifiers results across a wide range of classification tasks.
In addition, we observe the following:
(1) \knn{} can be \emph{on par with} neural network models, sometimes surpassing linear classifier counterparts. 
(2) Leveraging \knn{} is especially helpful for tasks beyond object classification and in low-data regimes.
These findings reveal that the
\knn{} classifier, which is ``rarely appropriate for use,''\footnote{From the \href{https://cs231n.github.io/classification/}{lecture notes} of Stanford CS231n Convolutional Neural Networks for Visual Recognition.} is still valuable in the deep learning era.
Our work makes the following key contributions:
\begin{enumerate}[leftmargin=*]
    \item A simple yet effective approach to augment standard neural network methods with \knn{} in both training and inference stages. We show that integrating \knn{} in both stages is more robust than integrating during test time only as proposed in previous work~\cite{solomatine2006eager,grave2017unbounded,orhan2018simple,khandelwal2020knnlm, khandelwal2020knnmt}.
    \item An extensive evaluation of the role of \knn{} in the context of a wide variety of classification tasks and experimental setups, demonstrating 
    the flexibility and generality of integrating \knn{}.
\end{enumerate}
We hope this work can broaden the landscape of computer vision and inspire future research that rethinks the role of old-school methods.

\section{Related Work}\label{sec:related}

\paragraph{\knn{} in the era of deep learning}
While \knn{} has declined in visibility relative to modern neural network based approaches in the last decade or so, it did not fall completely off the radar.
A \knn{} based model was shown to provide strong baselines for image captioning~\cite{devlin-etal-2015-language,devlin2015exploring}\footnote{\href{https://web.eecs.umich.edu/~justincj/slides/eecs498/FA2020/598_FA2020_lecture02.pdf}{Lectures} from University of Michigan EECS 498-007 / 598-005
Deep Learning for Computer Vision, which cover the same topics as CS231n, do, however, defend \knn{} with captioning as an example.},
image restoration~\cite{Ploetz2018NNN},
few-shot learning~\cite{wang2019simpleshot}, and evaluation of representation learning~\cite{wallace2020extending,caron2021dino}.
Nearest-Neighbor approaches are also used for improving interpretability and robustness of models against adversarial attacks~\cite{papernot2018deep,orhan2018simple,Junbo2019racnn}, language modeling~\cite{grave2017unbounded,khandelwal2020knnlm}, machine translation~\cite{tu-etal-2018-learning,khandelwal2020knnmt}, and learning a finer granularity than the one provided by training labels~\cite{touvron2021grafit}.
The notion of building associations and differentiating data instances has also been used in self-supervised learning~\cite{wu2018improving, wu2018unsupervised,chen2020simclr,he2020moco,chen2020mocov2,chen2021mocov3,caron2020swav,zbontar2021barlow}.
Most recently, Granot~\etal~\cite{granot2021drop} describe a patch-based nearest neighbors approach for single image generative models, producing more realistic global structure and less artifacts.
In this paper, we seek to explore how to augment existing neural network models with \knn{}, where the relative notion of similarity is based on the representations of the images.

\paragraph{Hybrid approach of eager and lazy learning}
Our work can be seen as a continuation of 
a line of work that combines eager learners with lazy learners~\cite{solomatine2006eager,grave2017unbounded,orhan2018simple,mi2020memory,khandelwal2020knnlm, khandelwal2020knnmt,leake2021supporting}. 
Eager learners, such as neural networks, are trained to provide a global approximating function that maps from input to output space.
Lazy learners, on the contrary, focus on approximating the neighborhoods around test examples~\cite{bontempi2001local}.
To leverage the ``best of both worlds,'' researchers have adopted simple interpolation strategies to combine these models.
The predictive results from \knn{} are linearly interpolated with those of an eager learner, either by a fixed scalar~\cite{solomatine2006eager,grave2017unbounded,orhan2018simple,khandelwal2020knnlm, khandelwal2020knnmt} or a learned weight~\cite{mi2020memory}.
Our present work takes a step further for fruitful pairings: in addition to the aforementioned approach, we propose to train a parametric classifier with the predictive results of \knn{} as guidance so the learned classifier will focus more on the examples misclassified by \knn{}.

\paragraph{Transfer learning}
Transfer learning~\cite{caruana1997multitask} has always been one of the ``tricks of the trade'' of machine learning practitioners. It has been widely used for evaluating representation learned under weakly-supervised~\cite{sun2017jft,yfcc100m,wslimageseccv2018,kolesnikov2019large,kolensnikov2020bit,li2021mopro,jia2021exploring}, semi-supervised~\cite{yalniz2019billion,yan2020clusterfit,xie2020noisy_student} and self-supervised learning paradigms~\cite{wu2018unsupervised,caron2018deepcluster,zhuang2019localagg,donahue2019bigbigan,ye2019e2e,caron2020swav,2020byol,he2020moco,misra2020pirl,chen2020simclr,li2021pcl,zbontar2021barlow,dang2021nearest}.
Transfer learning exploits the associations between the pre-trained task and the target task, especially when the data in the target task is scarce and high-capacity deep neural networks suffer from overfitting.
In this work, we utilize the transfer learning settings and use the visual representation learned from target task as input for \knn{} classifier. 

\begin{figure*}[ht]
\centering
\includegraphics[width=0.98\textwidth]{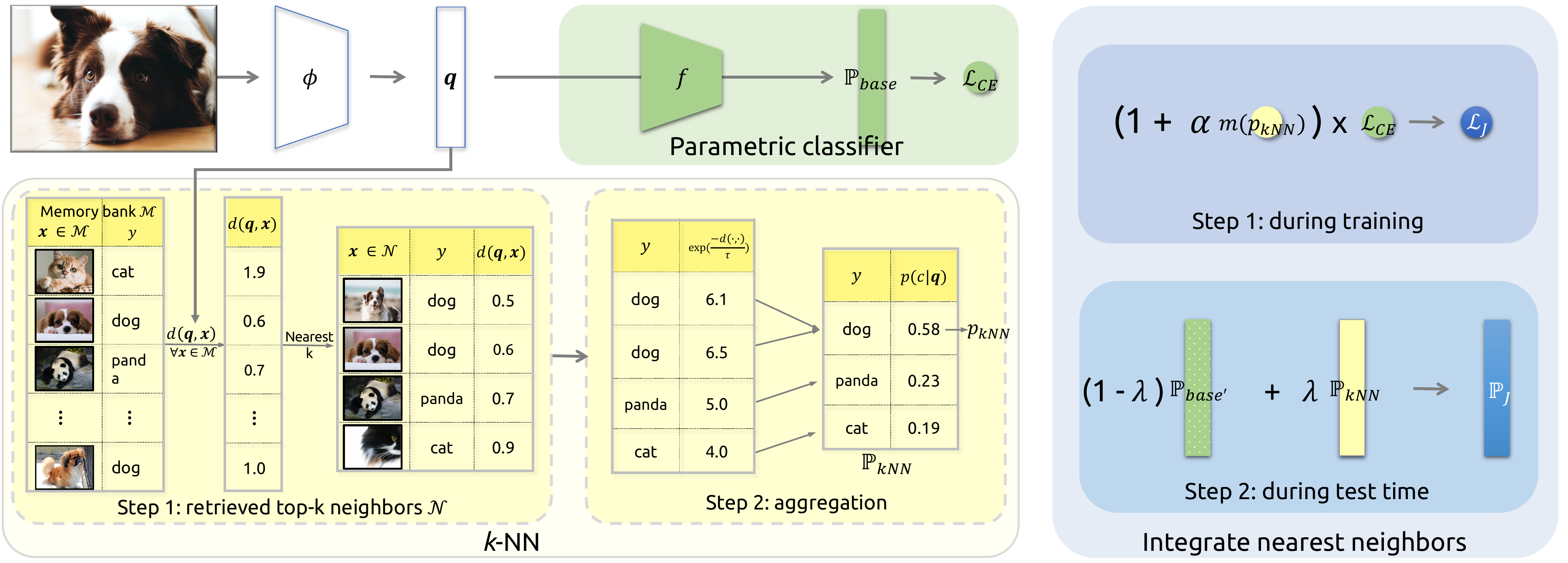}
\caption{
Method overview. Left: comparison between a parametric classifier and a \knn{} classifier. Right: Given an image, we integrate \knn{} classifier results to augment the conventional parametric classifier in two stages: 
(1): During training, we utilize \knn{} results as an indication of easy \vs~hard instances and rescale the loss accordingly;
(2): During test time, we linear interpolate results of two classifiers with a tunable parameter $\lambda$. Note we use the predicted results from the classifier learned using $\mathcal{L}_J$ from step 1.
See texts for more details.
}
\vspace{-0.3cm}
\label{fig:method}
\end{figure*}

\section{Approach}\label{sec:method}

Our goal is to investigate the role of the $k$-Nearest-Neighbor (\knn{}) classifier in the era of deep learning. 
Let $\phi$ be a visual feature extractor that transforms an input image $I$ into a numerical representation with dimensions $D$: $\vec{x}=\phi(I)  \in \R^{D}$.
$\vec{x}$ are used as input for \knn{}, which we describe in \cref{subsec:knn}.
We then introduce our method to integrate \knn{} with standard parametric classifiers in \cref{subsec:joint}. 
The overall framework is presented in \cref{fig:method}.

\subsection{Nearest neighbors revisited}
\label{subsec:knn}
$k$-Nearest-Neighbor (\knn{}) classifiers require no model to be fit.
Given a set of $n$ labeled images, we have the corresponding set of features $\mathcal{M} = \{\vec{x}_1, \dots, \vec{x}_n\}$, and a set of target labels $\{y_1, \dots, y_n\}$, $y\in [1, C]$.
Given a representation $\vec{q} = \phi(I_q)$ of a new image example $I_q$, 
\knn{} classifier can be described in the following two steps:

\paragraph{Step 1: Retrieve $k$ neighbors}
We first compute the Euclidean distance between the query $\vec{q}$ and all the examples in $\mathcal{M}$: $d(\vec{q}, \vec{x})$, $\forall \vec{x} \in \mathcal{M}$.
Then a set of $k$ representations are selected, which are the closest in distance to $\vec{q}$ in the embedded space. 
As commonly practiced in \knn{} (\eg,~\cite{friedman2017elements,wu2018unsupervised,wang2019simpleshot}),
we pre-process both $\vec{q}$ and features of $\mathcal{M}$ with $l2$-normalization and subtracting the mean across $\mathcal{M}$ ($\vec{\mu} = \frac{1}{n}\sum_{\vec{x} \in \mathcal{M}} \vec{x}$) from each feature.

\paragraph{Step 2: Aggregation}
Let $\mathcal{N}$ be the retrieved set of top-$k$ neighbors, and 
$\mathcal{N}_c$ be the subset of $\mathcal{N}$ where all examples have the same class $c$. 
The top-$k$ neighbors to $\vec{q}$ and their corresponding targets are converted into a distribution over $C$ classes. 
The probability of $\vec{q}$ being predicted as $c$ is:
\begin{equation}
    \label{eq:knn_prob}
    p_{k\text{NN}}(c \vert \vec{q}) = \frac{\sum_{\vec{x}\in \mathcal{N}_c}  \exp{} \left(-d(\vec{q}, \vec{x}) / \tau \right)}{\sum_{c \in C} \sum_{\vec{x}\in \mathcal{N}_c}  \exp{} \left(-d(\vec{q}, \vec{x}) / \tau \right)}\eqdot
\end{equation}
where $\tau$ is a temperature hyper-parameter. Higher $\tau$ produces a softer and more flattened probability distribution.

\paragraph{Comparing to parametric classifiers}
Given $\vec{q}$ and $C$ target classes, 
the probability of being predicted by a conventional parametric softmax classifier as target $c$ is
\begin{equation}
    \label{eq:base_prob}
    p(c \vert \vec{q}) = \frac{\exp{f_c(\vec{q})}}{\sum_{i=1}^C \exp{f_i(\vec{q})}}\eqcomma{}
\end{equation}
where $f_i(\cdot)$ is a parametric function that measures how well $\vec{q}$ matches the $i^{\text{th}}$ target class. 
In the example of a linear classifier $f(\vec{q}) = W\vec{q} + \vec{b}$.
The key difference to \knn{} is $\{W,\vec{b}\}$, which are parameters learned with a set of labeled training data.

\subsection{Integrating nearest neighbors}
\label{subsec:joint}

Leveraging the lazy learning benefits of \knn{}, we propose to augment the parametric classifier in both the training and test stages.

\paragraph{Step 1: Integrating during training}
Since \knn{}'s predictions given $\vec{q}$ can be easily computed, we wish to guide the network to attend to difficult examples during training.
In particular, we use \knn{} results to differentiate between easy \vs~hard examples.
Given a \knn{} probability of the ground-truth class $p_{k\text{NN}}$, we rescale the standard cross-entropy loss $\mathcal{L}_{CE}$ to adjust the relative loss for the misclassified or well-classified instances identified by \knn{}, respectively.
We employ the negative log-likelihood as the modulating factor $m(p_{k\text{NN}})$. 
The final loss $\mathcal{L}_J$ is defined as:
\begin{align}
    &m(p_{k\text{NN}}) = - \log{}(p_{k\text{NN}})\eqcomma \label{eq:modulate}\\
    &\mathcal{L}_J = \left(1 + \alpha m(p_{k\text{NN}}) \right)\mathcal{L}_{CE}\eqcomma
    \label{eq:joint}
\end{align}
$\alpha$ is a scalar to determine the contribution of each loss term.

Note we use training set as $\mathcal{M}$. Each example cannot retrieve itself as a neighbor. Thus $p_{k\text{NN}}$ is computed using the \emph{leave-one-out} distribution on the training set~\cite{wu2018improving}.
Furthermore, the idea of using a modulating factor to reweight loss of easy \vs~hard examples can be found in modern neural networks literature~\cite{focal_loss}.
Interestingly, we will show by experiments (\cref{sec:ana}) that our proposed loss is not sensitive to the initiation of modulating factor, indicating
that utilizing \knn{} predictions during training is the main reason for the observed improvements in \cref{sec:exp,sec:ana,sec:dis}.

\paragraph{Step 2: Integrating during test time}
Let $\mathbb{P}_{\text{\baseprime{}}}$ be the predicted class distribution produced by the trained classifier from Step 1, and $\mathbb{P}_{k\text{NN}}$ be that of a \knn{} classifier.
We calculate the final probability distribution $\mathbb{P}_J$ as:
\begin{equation}
\label{eq:jointinf}
   \mathbb{P}_J = \lambda \mathbb{P}_{k\text{NN}} + ( 1 - \lambda) \mathbb{P}_{\text{\baseprime{}}}\eqcomma
\end{equation}
where $\lambda\in [0, 1]$ is a tuned hyper-parameter.
Since the \knn{} distribution assigns non-zero probability to at most $k$ classes, \cref{eq:jointinf} also ensures non-zero probability over the entire classes.

\begin{figure*}[t]
\centering
\includegraphics[width=0.95\textwidth]{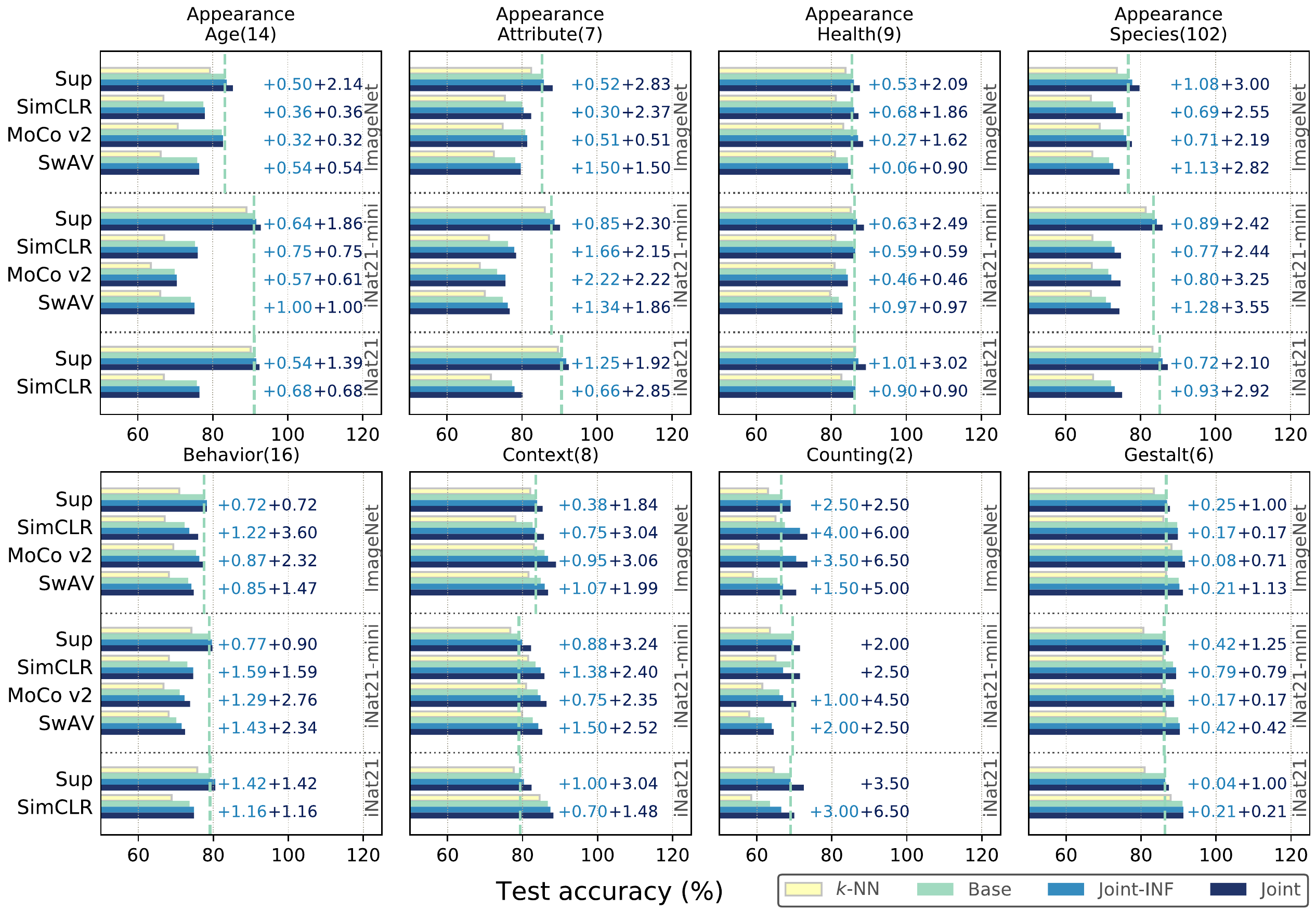}
\vspace{-0.2cm}
\caption{
Results on \newt{}. Every experiment uses \rn{}-50 as pre-training backbone. 
The y-axis denotes the pre-training objectives while the text on the right specify the pre-training datasets. 
Performance gains of \jointinf{} and \joint{} for each representation are presented if applicable.
The green dashed line in each subplot represent the \base{} results with supervised feature extractors.
}
\vspace{-0.2cm}
\label{fig:newt_results}
\end{figure*}

\section{Experiments}\label{sec:exp}

We conduct extensive experiments using three sets of classification tasks with benchmarking datasets,
comparing the performance of methods that incorporate \knn{}, with conventional parametric classifier and $k$-nearest neighbor classifier.
We first describe our experimental setup (\cref{subsec:imp_details}), and demonstrate the effectiveness of our integrated methods on natural world binary classification~\cite{VanHorn2021newt} (\cref{subsec:newt_exp}), 
fine-grained object classification~\cite{WahCUB_200_2011,nilsback2008automated,parkhi2012dogs,gebru2017cars} (\cref{subsec:fgvc_exp}) and \imagenet{} classification~\cite{imagenet_cvpr09} (\cref{subsec:imgnet_exp}).

\subsection{Evaluation protocols and details} 
\label{subsec:imp_details}

We adopt linear evaluation as it is a common feature representation evaluation protocol~\cite{wslimageseccv2018,goyal2019scaling,misra2020pirl,he2020moco}.
Pre-trained models are used as visual feature extractors, where the weights of the image encoders are fixed\footnote{
See the~\cref{subsec:finetune_supp} for fine-tuning results for completeness, where all the parameters of the pre-trained models are fine-tuned in an end-to-end manner for downstream tasks.
}. 
We compare the proposed method (denote as \joint{}) with the following approaches:
(1) \base{}: which is the vanilla linear classifier, whose performance usually indicates how effective the learned representations are;
(2) \knn{} described in~\cref{subsec:knn};
(3) \jointinf{}: which simply linear interpolate \knn{} and \base{} result during test time.

We experiment with a wide range of feature representations, categorized by its (1) pre-trained objectives, (2) pre-trained datasets, and (3) feature backbones.
We consider 8 different training objectives including \suplong{} (\supshort{}) setting and self-supervised objectives:
\simclr{}~\cite{chen2020simclr}, 
\mocotwo{}~\cite{he2020moco,chen2020mocov2},
\moco{}~\cite{chen2021mocov3},
\swav{}~\cite{caron2020swav},
\btwins{}~\cite{zbontar2021barlow},
\dino{}~\cite{caron2021dino},
\moby{}~\cite{xie2021moby}. 
The pre-training datasets can be either \imagenet{}~\cite{imagenet_cvpr09} or \inatlong{} (\inat{})~\cite{VanHorn2021newt}.
The feature backbones include \rn{}~\cite{he2016rn}, Vision Transformers (\vit{})~\cite{dosovitskiy2020vit} and Swin Transformer (\swin{}~\cite{liu2021swin}).
\cref{supsec:detail} provides more implementation details, including 
an itemized list of the feature representations configurations, dataset specifications, and a full list of hyperparameters used (batch sizes, learning rates, decay schedules, \etc).

\begin{figure*}[t]
\centering
\includegraphics[width=0.95\textwidth]{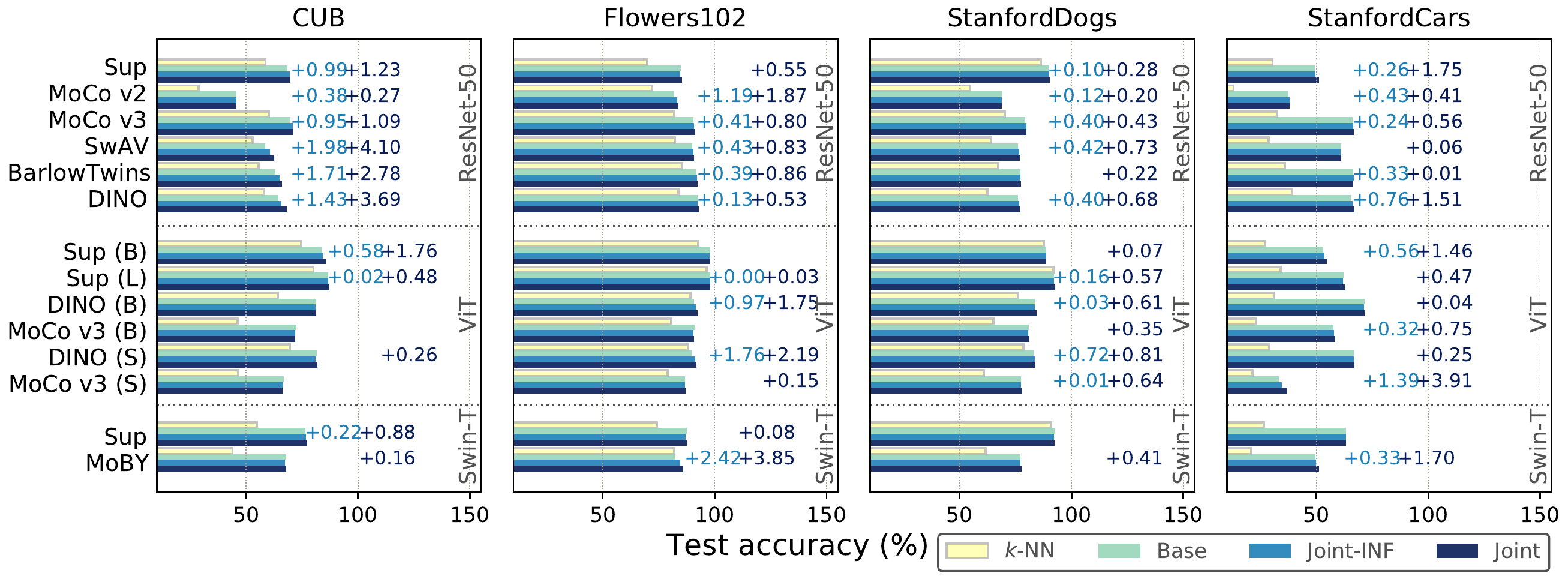}
\vspace{-0.2cm}
\caption{
Results on fine-grained down-stream datasets. All features are pre-trained on \imagenet{}. The y-axis denotes the pre-training objectives while the text on the right specify the feature extractor backbones. We also include the performance gains of \joint{} and \jointinf{} for each representation evaluated if applicable.
}
\vspace{-0.3cm}
\label{fig:fgvc_results}
\end{figure*}
\begin{figure}
    \centering
    \includegraphics[width=0.9\columnwidth]{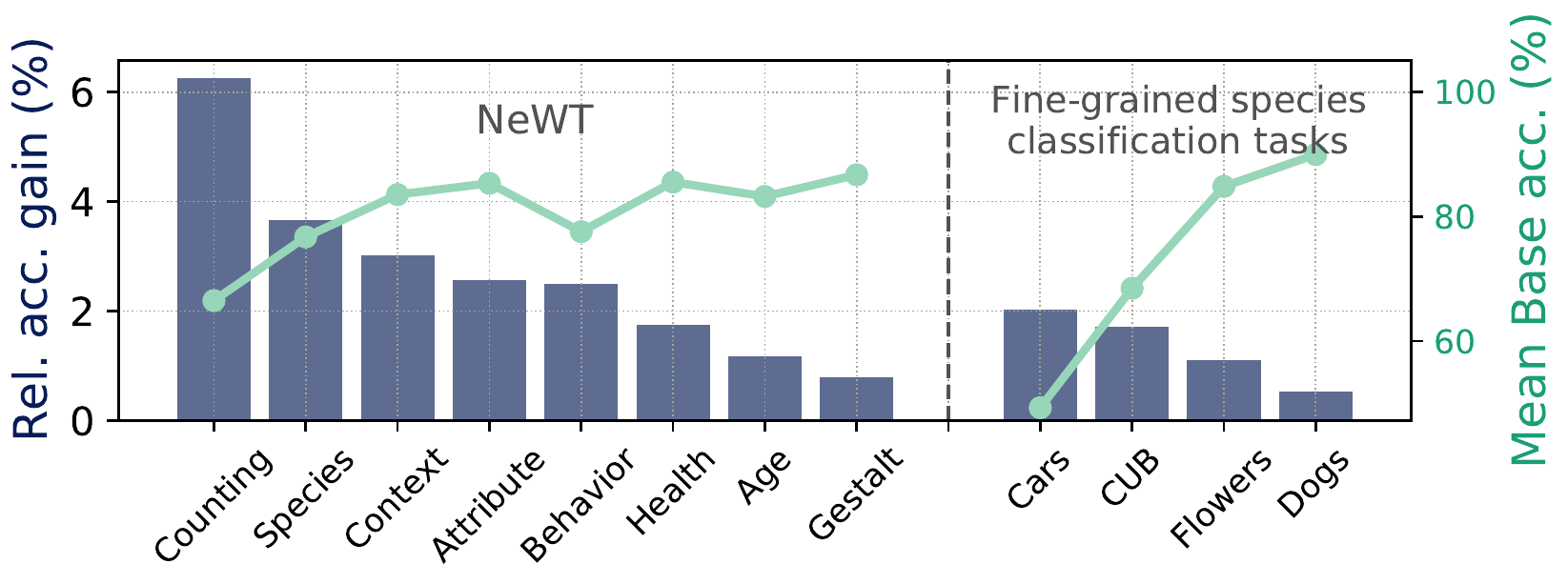}
    \vspace{-0.2cm}
    \caption{Relative average accuracy gain ($\%$) \vs~average \base{} accuracy rate ($\%$) across all the evaluated features for \newt{} and fine-grained object classification tasks.}
    \vspace{-0.4cm}
    \label{fig:rel_acc}
\end{figure}

\subsection{Natural world binary classification}
\label{subsec:newt_exp}

\paragraph{Settings} We first evaluate the effect of integrating \knn{} on a collection of Natural World Tasks (\newt{})~\cite{VanHorn2021newt}. It comprises 164 binary classification tasks related to different aspects of natural world images, including appearances, behavior, context, counting, and gestalt. Following~\cite{VanHorn2021newt}, we categorize the tasks into 8 coarse groups and report the mean test accuracy score of each group.
In addition to features pre-trained on \imagenet{} and \inat{}, 
we also report the features pre-trained on a randomly sampled subset (18.6 $\%$) of \inat{}, denoted as \inat{}-mini.
\cref{fig:newt_results} summarizes the experimental results. \cref{subsec:results_supp} includes the accuracy values for all tasks.

\paragraph{Results}
\cref{fig:newt_results} shows that
\joint{} noticeably surpasses the vanilla linear classifier counterpart (\base{}) for all 8 groups of \newt{} and all the feature representations evaluated.
In general, relative accuracy gains and \base{} results are negatively correlated. \cref{fig:rel_acc} summarizes this trend.
For example,
we observe non-trivial performance gains of both \jointinf{} and \joint{} on the \emph{counting} task: +5.2$\%$ / +9.7$\%$ relative gains over base{} (67.0$\%$) for \jointinf{} (70.5$\%$) and \joint{} (73.5$\%$), respectively, using supervised pre-training methods on \inat{}.
This task group receives the lowest mean \base{} accuracy across all features (66.5$\%$) among \newt{},
suggesting \joint{} is more effective on the poorly performing parametric classifiers.

\subsection{Fine-grained object classification}
\label{subsec:fgvc_exp}

\paragraph{Settings}
We further demonstrate the effect of \joint{} by benchmarking fine-grained object classification tasks: \longcub{} (\cub{}~\cite{WahCUB_200_2011}), \flowers{}~\cite{nilsback2008automated},
\dogs{}~\cite{parkhi2012dogs},
\cars{}~\cite{gebru2017cars}.
All four datasets are multi-class classification tasks. If a certain dataset only has \train{} and \test{} set publicly available, we randomly split the training set into \train{} (90\%) and \val{} (10\%) and optimize the top-$1$ accuracy scores on \val{} to select hyperparameters.
\cref{suppsec:tasks} includes additional information about these tasks. 
\cref{fig:fgvc_results} presents the test accuracy rate for the \test{} set of these datasets.
All feature representations are pre-trained on \imagenet{}.

\paragraph{Results}
From~\cref{fig:fgvc_results}, we can see that:
(1) \joint{} offers better or comparable results than the linear classifier counterpart. 
(2) \joint{} is more generalizable than \jointinf{}. For \rn-50 backbone, \joint{} achieves accuracy gains over all the features across four datasets, while \jointinf{} brings reduced improvement over a subset of the evaluated features.
We observe that \jointinf{} sometimes can be prone to overfitting, where \val{} results are higher yet \test{} results are not.
(3)  Comparing to \newt{}, we see a similar trend where the improvement of \joint{} is negatively correlated with \base{} results. See~\cref{fig:rel_acc} (right).
(4) \joint{} achieves higher accuracy gains on \rn{} backbone in general, comparing with other transformer-based backbones (\vit{} and \swin{}).
This could suggest that the transferability of \rn{} features is not fully exploited using a linear classifier alone.

\begin{figure*}[t]
\begin{subfigure}{.33\textwidth}
  \centering
  \includegraphics[scale=0.4]{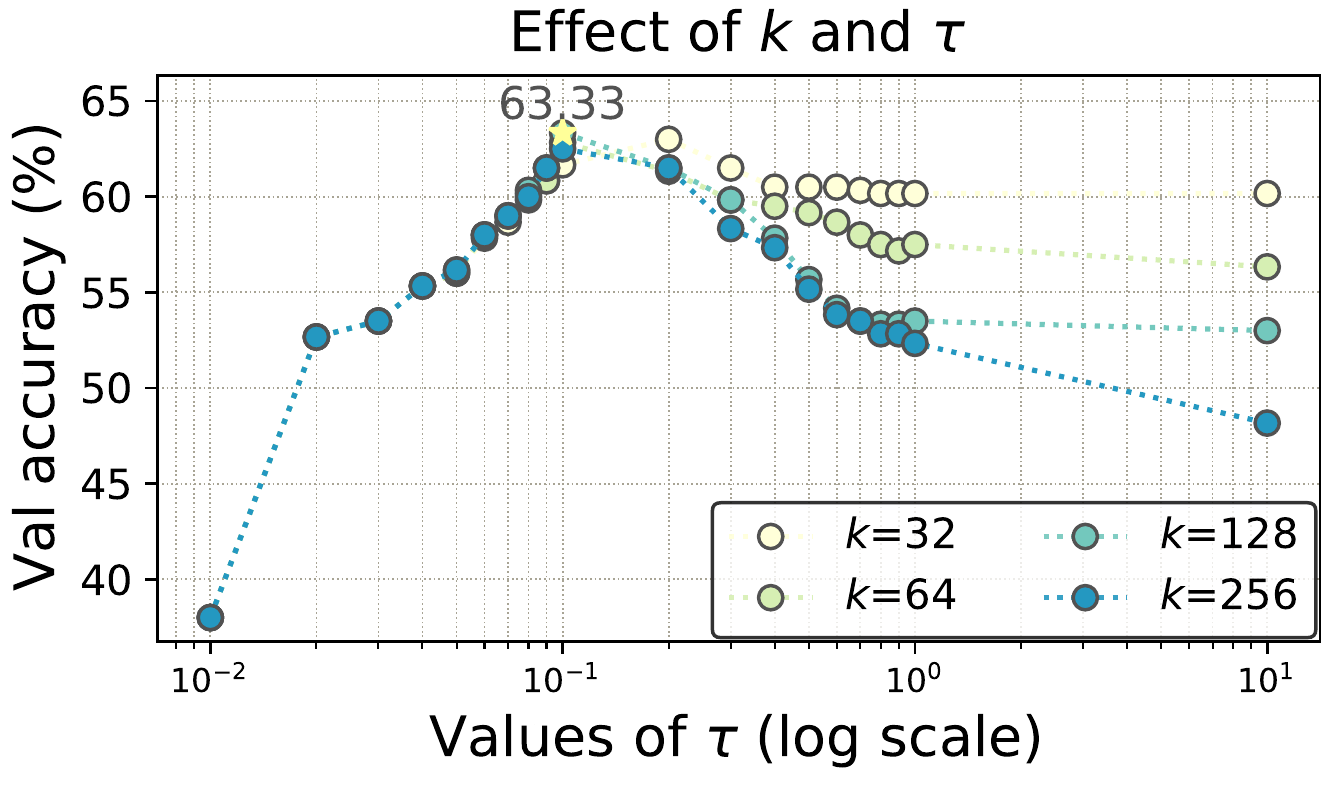}
  \caption{Effect of $\tau$.}
  \label{fig:ablation_kt1}
\end{subfigure}%
\begin{subfigure}{.33\textwidth}
  \centering
  \includegraphics[scale=0.4]{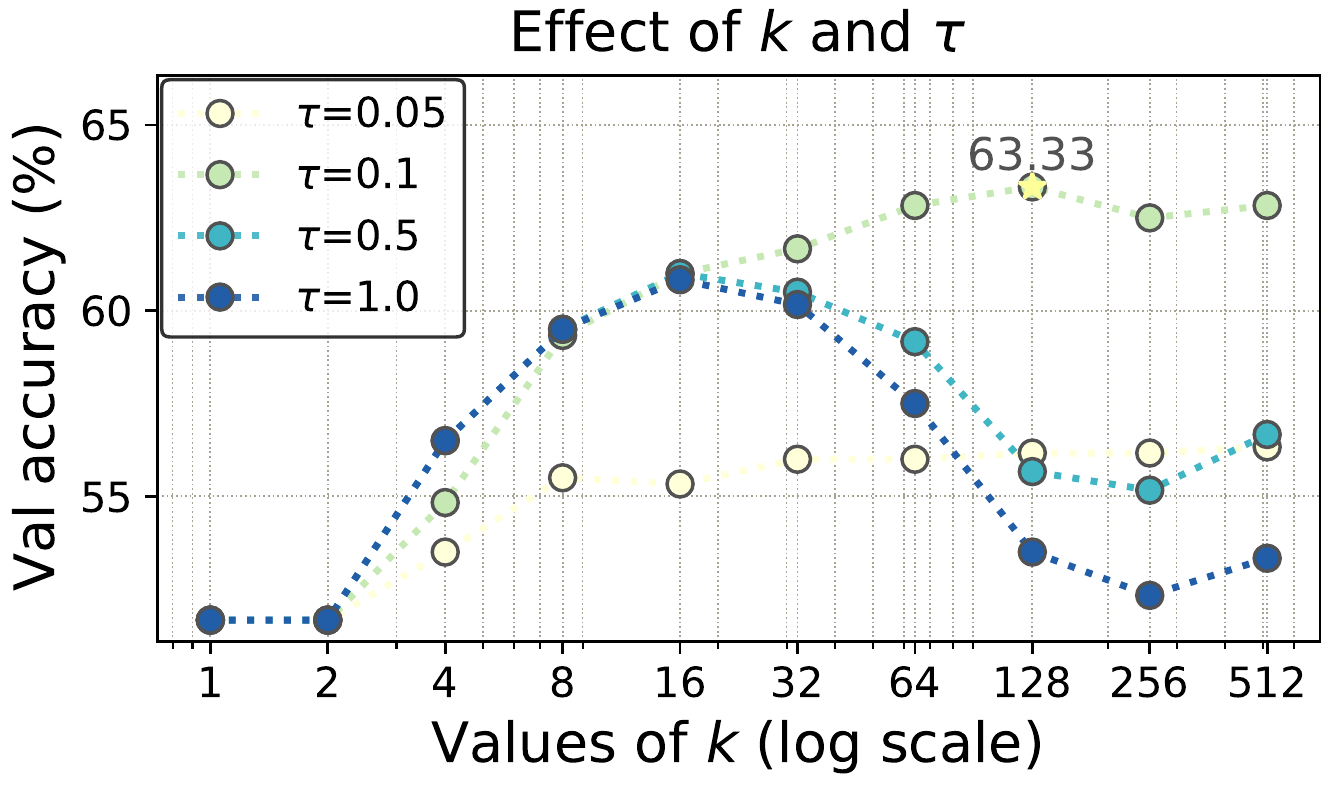}
  \caption{Effect of $k$.}
  \label{fig:ablation_kt2}
\end{subfigure}
\begin{subfigure}{.33\textwidth}
  \centering
  \includegraphics[scale=0.4]{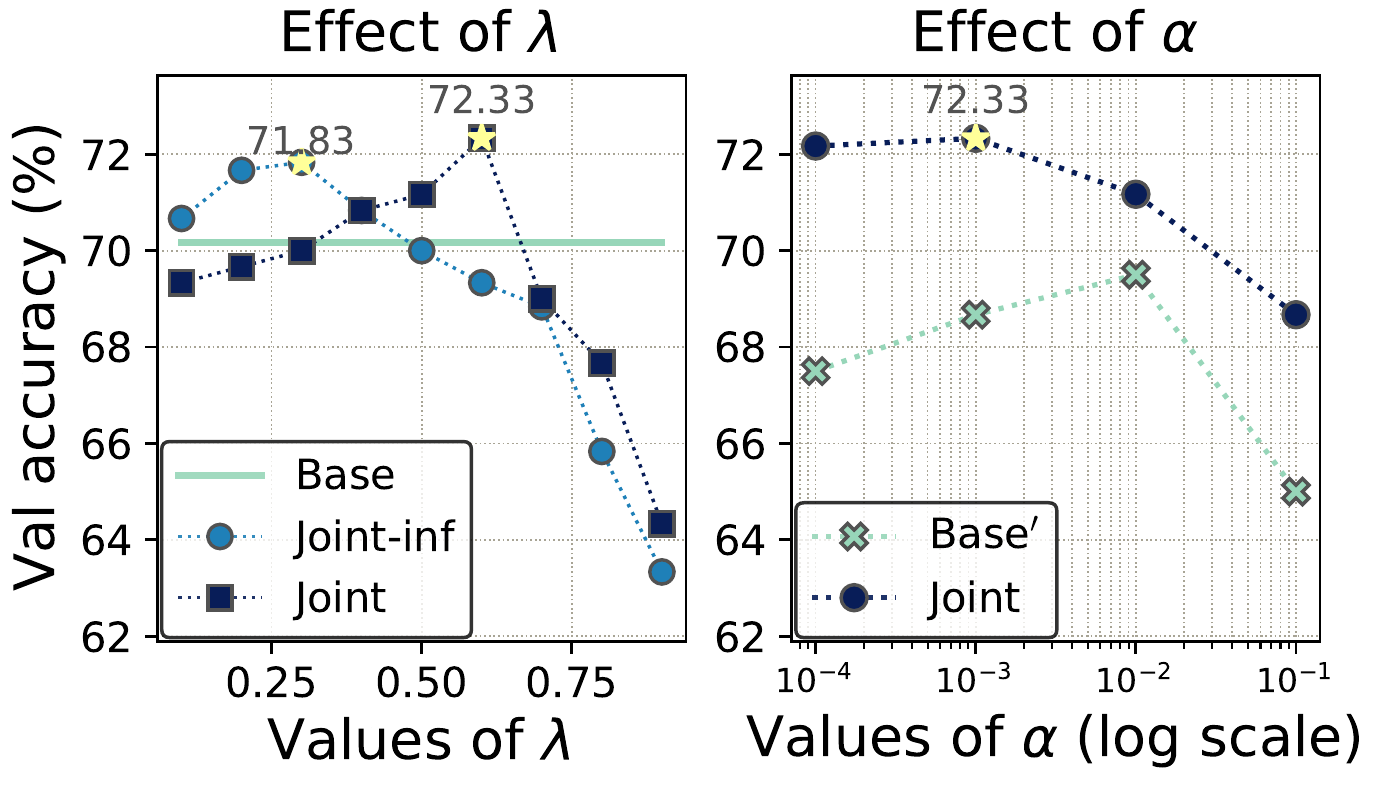}
  \caption{Effect of $\lambda$ and $\alpha$.}
  \label{fig:ablation_lambdaalpha}
\end{subfigure}
\vspace{-0.3cm}
\caption{Tuning \knn{} and \jointinf{} and \joint{}.
(a,b): Effect of $k,\tau$ on the \knn{} accuracy scores. 
(c): Effect of interpolating factor $\lambda$ on \jointinf{} and $\alpha$ on \joint{} accuracy scores.
All results are for \cub{}~\val{} dataset with supervised imageNet pre-training ResNet-50 backbone. 
}
\vspace{-0.6cm}
\label{fig:hp_ablation}
\end{figure*}

\subsection{ImageNet classification}
\label{subsec:imgnet_exp}

\paragraph{Settings} We evaluate the image classification task using \imagenet{} (ILSVRC 2012)~\cite{imagenet_cvpr09}, which contains 1.3 million training and 50k validation natural scene images with 1000 classes. 
For supervised objectives, we freeze the feature extractors and re-initiate the last layer for the \joint{} approach.
\cref{tab:imgnet_results} summarizes the results with top-1 accuracy rate.

\begin{table}
\scriptsize
\begin{center}
\resizebox{\columnwidth}{!}{%
\begin{tabular}{l l rrrr}
\toprule

\multirow{2}{*}{\textbf{Backbone}} 
&\multirow{2}{*}{\shortstack[l]{\textbf{Pre-trained}\\\textbf{Objective}}} 
&\multicolumn{4}{ c }{\textbf{Top 1 Accuracy ($\%$)}} \\
\cmidrule{3-6}

&& \textbf{\knn{}}  &\textbf{\base{}} &\textbf{\jointinf{}} &\textbf{\joint{}}
\\
\midrule
\multirow{5}{*}{\rn{}-50}
&\suplong{} &72.65 &76.01 &\Rise{0.37}76.39 &\Rise{0.50}\textbf{76.52}
\\
&\moco{}~\cite{chen2021mocov3} &68.51  &73.88 	&\Rise{1.25}75.13 &\Rise{1.28}\textbf{75.16}
\\
&\swav{}~\cite{caron2020swav}	&68.54	&$^\star$74.63	&\Rise{1.48}\textbf{76.11}	&\Rise{1.40}76.04 
\\
&\btwins{}~\cite{zbontar2021barlow}	&67.85	&$^\star$72.88	&\Rise{1.52}\textbf{74.40}	&\Rise{1.31}74.18 
\\
&\dino{}~\cite{caron2021dino}	&68.40	&$^\star$74.38	&\Rise{1.60}\textbf{75.97}	&\Rise{1.54}75.91
\\
\midrule

\vit{}-B/16 &\suplong{}~\cite{dosovitskiy2020vit} &77.27 &75.70	&\Rise{1.78}\textbf{77.48} &\Rise{2.62}78.32  
\\
\vit{}-L/16	&\suplong{}~\cite{dosovitskiy2020vit} &80.25	&79.20	&\Rise{1.21}80.41 &\Rise{1.61}\textbf{80.81} 
\\
\vit{}-S/16 &\moco{}~\cite{chen2021mocov3} &69.01
&$^\star$73.33 	&\Rise{1.25}\textbf{74.58} &\Rise{1.23}74.56
\\

\vit{}-B/16 &\moco{}~\cite{chen2021mocov3} &72.24 &$^\star$76.35 	&\Rise{1.02}77.38 &\Rise{1.05}\textbf{77.41} \\

\midrule

\multirow{2}{*}{\swin{}-T} 
&\suplong{}~\cite{liu2021swin} &80.06	&$^\star$80.98	&\Rise{0.25}\textbf{81.23} &\Rise{0.25}\textbf{81.23} \\
&\moby{}~\cite{xie2021moby} &68.87	&$^\star$74.68	&\Rise{0.84}75.52	&\Rise{0.91}\textbf{75.59}
\\
\bottomrule
\end{tabular}
}
\vspace{-0.2cm}
\caption{Results on \imagenet{}. All features are pre-trained on \imagenet{}. \base{} results with $^\star$ are run by us. 
\textcolor{green_im}{($\cdot$)} indicates the difference to \base{}.
Methods that incorporate nearest neighbors consistently outperform the \base{} and \knn{} counterparts.}
\label{tab:imgnet_results}
\vspace{-0.5cm}
\end{center}
\end{table}

\paragraph{Results}
Under both protocols (\jointinf{} and \joint{}), integrating nearest neighbors achieves consistently better performance for \imagenet{} than \base{} across three backbone choices and all the pre-training objectives.
Comparing with \joint{} results, we observe that \jointinf{} offers comparable performance, sometimes even outperforms \joint{} by a slight margin (0.06-0.21) for \imagenet{} experiments. This is different from the previous two experiments in~\cref{subsec:newt_exp,subsec:fgvc_exp}, suggesting that \joint{} is more suitable for downstream tasks that possibly have larger domain shift than the pre-trained dataset.
Note that we use in-house \base{} results for fair-comparison with \joint{}.
Some \base{} results are different (within 1$\%$) from the reported results from prior work. \cref{subsec:results_supp} includes experiments with the released linear classifiers' weights.

\section{Analysis}\label{sec:ana}

To better understand the proposed integration approach, we conduct ablation studies and qualitative analysis using \cub{}. 
We define \baseprime{} as the classifier trained with $\mathcal{L}_J$ of~\cref{eq:joint} without test time integration.
More results and analysis are included in the~\cref{supsec:result}.

\begin{figure*}[t]
\centering
\includegraphics[width=\textwidth]{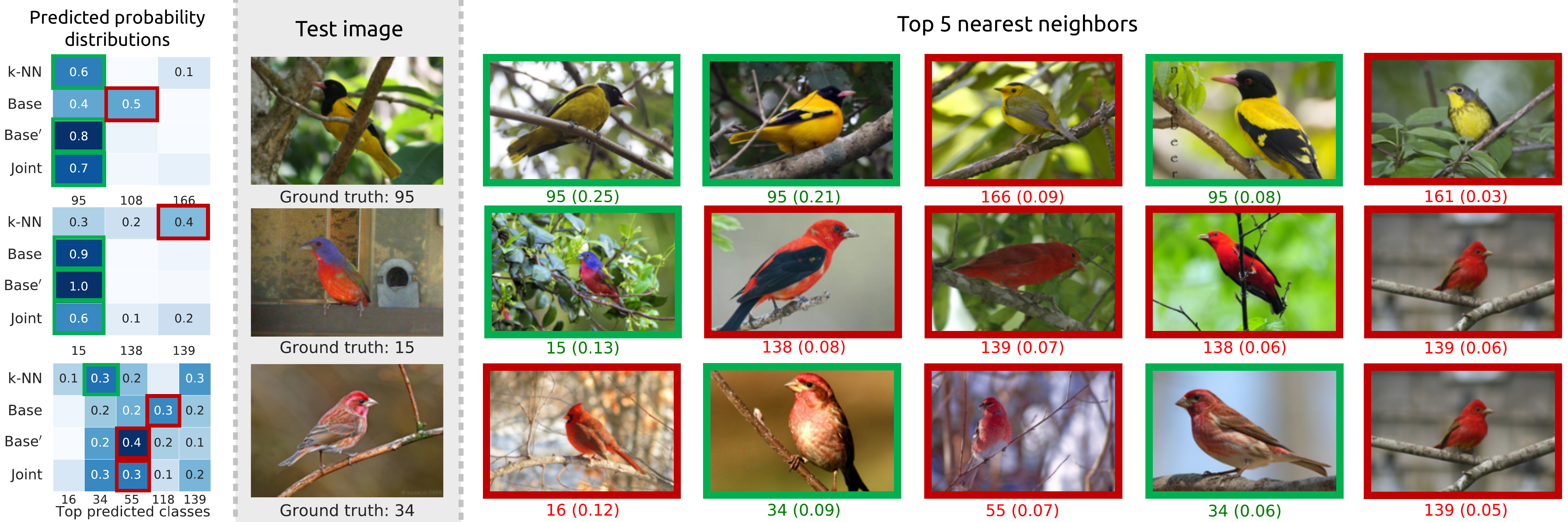}
\vspace{-0.6cm}
\caption{
Qualitative results where \knn{} and \base{} produce different predictions. We show the predicted probability distribution (left), the test image (middle), and the top-5 nearest neighbors from the training set on the right. 
Only predicted probabilities $>0.1$ are presented in the left due to space constraint. 
Title of each retrieved samples follows the format: \texttt{class index(} $\exp{\left(-d(\cdot, \cdot) / \tau \right)}$ from~\cref{eq:knn_prob} \texttt{)}.
}
\vspace{-0.3cm}
\label{fig:qual}
\end{figure*}

\paragraph{Modulating factor initiations}
In previous experiments, we use negative log-likelihood (NLL) as a function of \knn{} predicted probability. The modulating factor is used to weight samples based on \knn{} prediction quality during training. 
A relatively modern method (comparing to \knn{} which was first proposed in 1951~\cite{fix1985knn}) to instance-based weighting is focal loss~\cite{focal_loss}, which proposes to weight training samples with $(1 - p)^{\gamma}$, where $p$ is the ground-truth class probability produced by a parametric classifier.
Like NLL, this initiation also reduces the relative loss for well-classified examples. See~\cref{supsec:result} for visualization and further analysis of both modulating factor choices.

We thus adopt focal loss modulating factor as an alternative initiation for \joint{}: $(1 - p_{k\text{NN}})^{\gamma}$.
We train \joint{} with focal weighting strategy (\joint{}-Focal) using identical settings as before on \cub{} and report the results in~\cref{tab:focal}. 
\joint{}-Focal achieves nearly the same accuracy as those trained with negative log-likelihood (\joint{}-NLL).
Both weighting strategies for \joint{} improve vanilla linear classifier performance by up to 7$\%$, suggesting that the exact form of modulating factor is not crucial.
A similar finding was also discussed in~\cite{focal_loss}. \joint{}-Focal is a reasonable alternative for \joint{}-NLL and works well in practice.

\begin{table}
\scriptsize
\begin{center}
\resizebox{\columnwidth}{!}{%
\begin{tabular}{l  rrr rr}
\toprule
\multirow{2}{*}{\shortstack[l]{\textbf{Pre-trained}\\\textbf{Objective}}} 
&\multicolumn{3}{ c }{\textbf{Top 1 Accuracy ($\%$)}} 
\\
\cmidrule{2-4}
& \textbf{\base{}} &\textbf{\joint{}-NLL} &\textbf{\joint{}-Focal} 
&$\alpha$  &$\gamma$
\\
\midrule

\suplong{} 
&68.48	&\Rise{1.23}69.71	&\Rise{1.47}\textbf{69.95}	&0.001 &2 
\\
\moco{}~\cite{chen2021mocov3} 
&69.76	&\Rise{1.09}\textbf{70.85}	&\Rise{1.04}70.80	&0.0001	&0.5
\\
\swav{}~\cite{caron2020swav}	
&58.53	&\Rise{4.10}\textbf{62.63}	&\Rise{3.91}62.44	&0.01	&0.5
\\
\btwins{}~\cite{zbontar2021barlow}	
&63.12	&\Rise{2.78}65.90	&\Rise{2.86}\textbf{65.98}	&0.001	&2
\\
\dino{}~\cite{caron2021dino}
&64.41	&\Rise{3.69}\textbf{68.10}	&\Rise{3.61}68.02	&0.01	&0.5
\\
\bottomrule
\end{tabular}
}
\vspace{-0.2cm}
\caption{Negative log-likelihood \vs focal loss weight factors comparison. All features are pre-trained on \imagenet{} with \rn-50{}.
\textcolor{green_im}{($\cdot$)} indicates the difference to \base{}.
The differences between both weighting schemes are within 0.35$\%$ of each other. 
}
\label{tab:focal}
\vspace{-0.5cm}
\end{center}
\end{table}

\paragraph{Tuning \knn{}}
\cref{fig:ablation_kt1,fig:ablation_kt2} present the effect of number of nearest neighbors considered ($k$) and temperature value ($\tau$ from~\cref{eq:knn_prob}) on the \val{} accuracy score of \cub{}. 
\cref{fig:ablation_kt1} shows that \knn{} achieves better performance as $\tau$ increases up to $0.1$. 
$\tau=0.01$ produces over-concentrated distribution and makes the model overfit to the most similar retrievals.
$\tau > 0.1$ can deteriorate the performance through over-smoothing the probability distribution and increasing the entropy~\cite{wang2021understanding}. 
\cref{fig:ablation_kt2} shows that for a fixed temperature value, the performance does not always improve using a larger $k$ since more nearest neighbors retrieved will likely include more instances with wrong labels. 
Combinations of larger values of $\tau$ (\eg $\tau > 0.1$) and larger $k$ ($k \geq 16$) will lead to even further reduced \knn{} accuracy rates.
The effect of $k$ diminishes gradually as $\tau$ decreases (\eg $\tau = 0.05$).

\paragraph{Tuning \joint{}}
To investigate the effect of how much the model relies on \knn{},
we also ablate the linear interpolation parameter $\lambda$ in~\cref{eq:jointinf} and the scaling factor $\alpha$ in~\cref{eq:joint}.
\cref{fig:ablation_lambdaalpha} summarizes the results.
\cref{fig:ablation_lambdaalpha} (left) shows that the optimal $\lambda$ increases for \joint{} comparing to \jointinf{}, as the model relies more on \knn{} with a specific weighting loss during training.
\cref{fig:ablation_lambdaalpha} (right) shows that the optimal values of $\alpha$ are different between the linear classifier learned under \joint{} framework (\baseprime{}) and \joint{} ($0.001$ and 0.01, respectively), 
highlighting the fact that the linear classifier that is complementary to \knn{} is not always the best performing classifier.

\paragraph{Qualitative Analysis}
To further understand the role of \knn{} and the benefit of \joint{},
we manually inspect cases in which \base{} and \knn{} produce different predictions with $\phi(\cdot)$ being the supervised pre-training on \imagenet{}.
\cref{fig:qual} shows such examples, along with others in the Appendix.
For each test example, we plot the predicted probability distributions of \joint{} and other approaches (\knn{}, \base{}, \baseprime{} which is the linear classifier trained with \knn{} weighted loss in~\cref{eq:joint}).
Top 5 nearest neighbors from the \train{} split of \cub{} are also visualized at the right hand side of the figure.
\knn{} can produce correct predictions
when \base{} fails, especially when \base{} produces high entropy distributions (top and bottom cases). 
Via reweighting the easy \vs~hard examples identified by the \knn{} results, \baseprime{} 
produces a larger probability for the correct class in all three cases. 
However, the bottom row shows that \joint{} and \baseprime{} produces incorrect result even when \knn{} is correct, possibly due to the existence of 5 fine-grained classes. The visual representation (input for \knn{}) seems not discriminative enough to distinguish the subtle differences among these classes.

\begin{figure}
\centering
\includegraphics[width=\columnwidth]{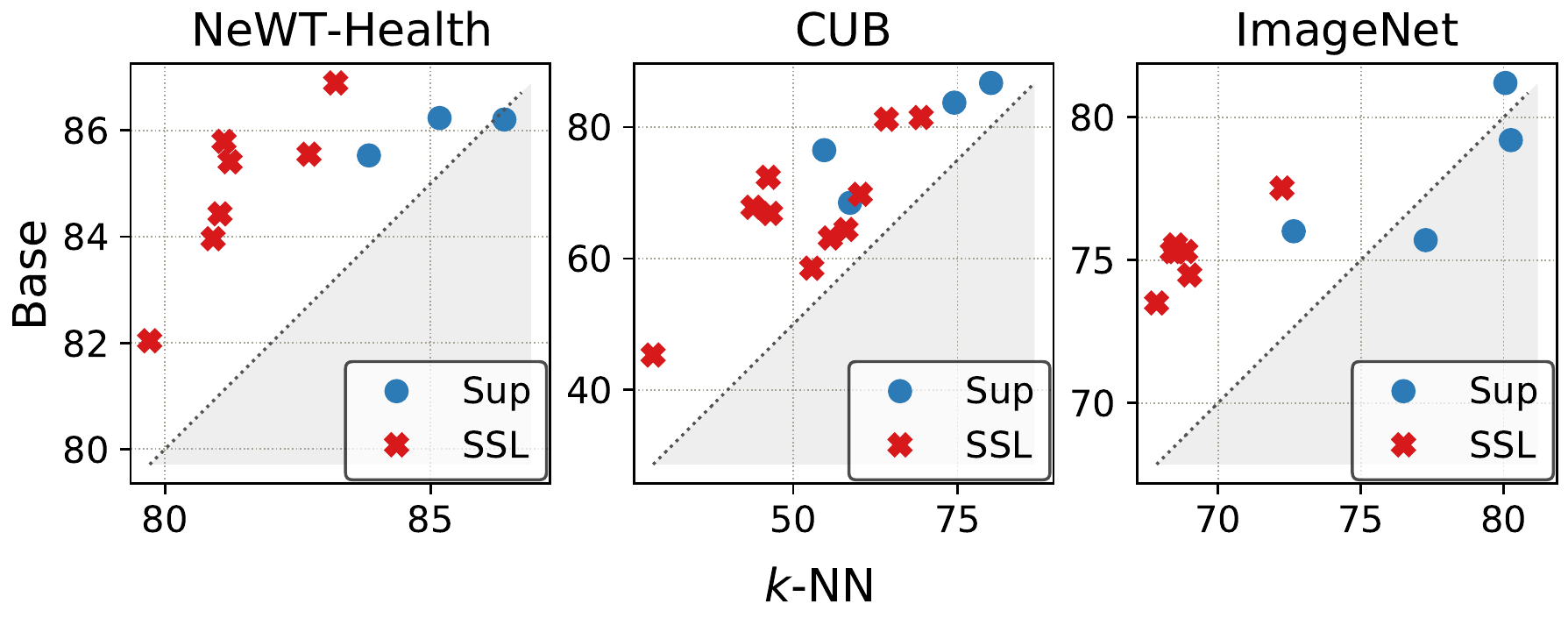}
\vspace{-0.8cm}
\caption{Supervised (Sup) \vs~self-supervised (SSL) features perform differently between \knn{} (x-axis) and \base{} (y-axis).  Each point represents accuracy rate of one feature configuration from~\cref{fig:newt_results,fig:fgvc_results}, and~\cref{tab:imgnet_results}.
}
\vspace{-0.4cm}
\label{fig:knn_vs_base}
\end{figure}

\vspace{-0.2cm}
\section{Discussion}\label{sec:dis}

\paragraph{Rethinking \knn{} and visual representations}
Our experiments show that in addition to features learned with a self-supervised \vit{} backbone, visual representations of diverse specifications (pre-training objectives, dataset, and backbones) can produce competitive \knn{} results.
This observation expands the findings of Caron \etal~\cite{caron2021dino}.
In particular, the relative accuracy between \base{} and \knn{} varies across pre-training objectives; see~\cref{fig:knn_vs_base}. The supervised pre-training features are more favorable for \knn{} in general. The performance gap between \knn{} and \base{} is smaller for these supervised features.

\begin{figure*}
\centering
\includegraphics[width=0.98\textwidth]{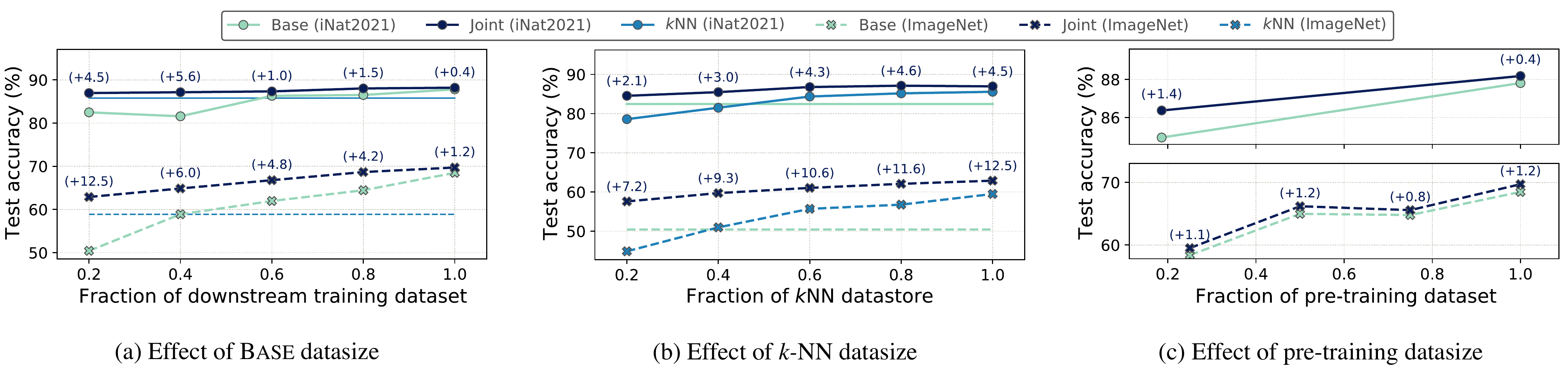}
\vspace{-0.2cm}
\caption{Dataset size ablation results with \cub{}. 
We study how the amount of data used for (a) \base{}, (b) \knn{}, and (c) pre-training process affects the performance. 
We use full \knn{} data for (a), 20$\%$ \base{} data for (b), full \knn{} and \base{} data for (c). (a-b) use full pre-training data.
All experiments use \rn{}-50 as backbone. Legend follows format: \texttt{method (pre-training data)}.
($\cdot$) denotes the differences between \joint{} and \base{} under each variation.
}
\vspace{-0.3cm}
\label{fig:ablation_size}
\end{figure*}

\begin{figure*}
\begin{subfigure}{.33\textwidth}
  \centering
  \includegraphics[scale=0.4]{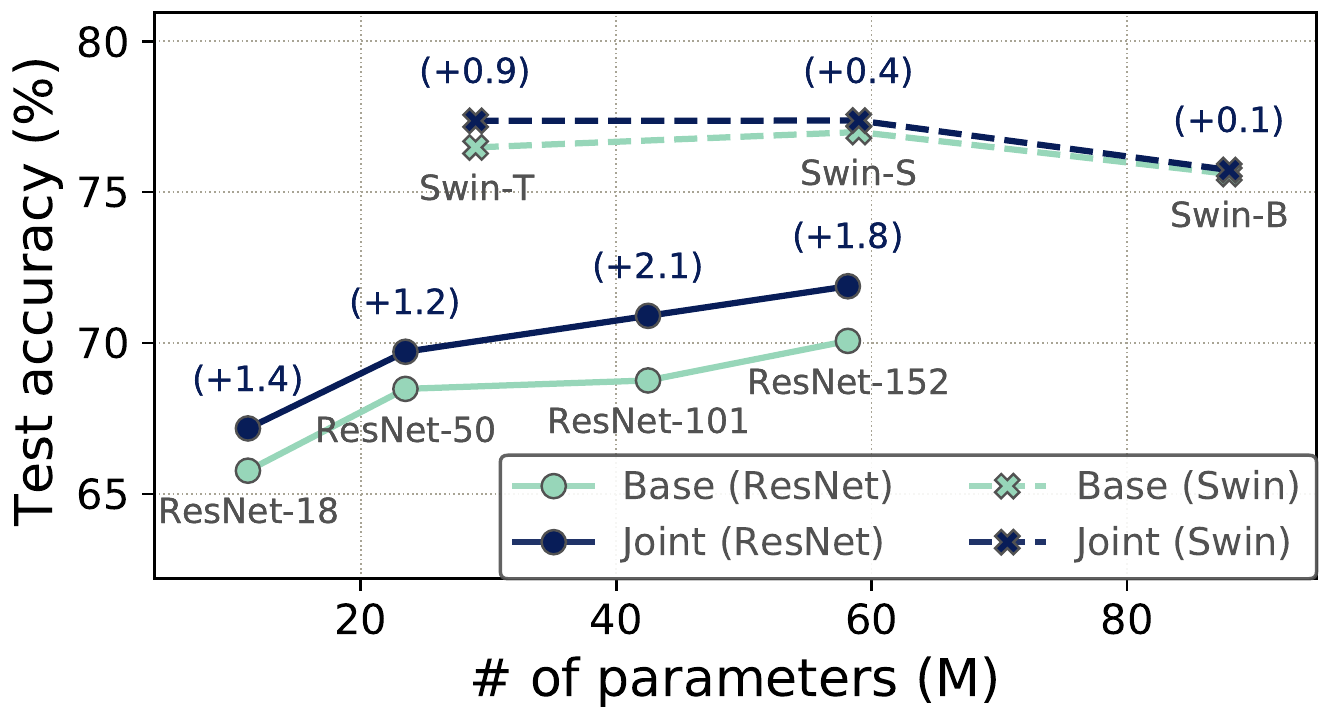}
  \caption{Effect of pre-training backbone.}
  \label{fig:ablation_backbone}
\end{subfigure}%
\begin{subfigure}{.33\textwidth}
  \centering
  \includegraphics[scale=0.4]{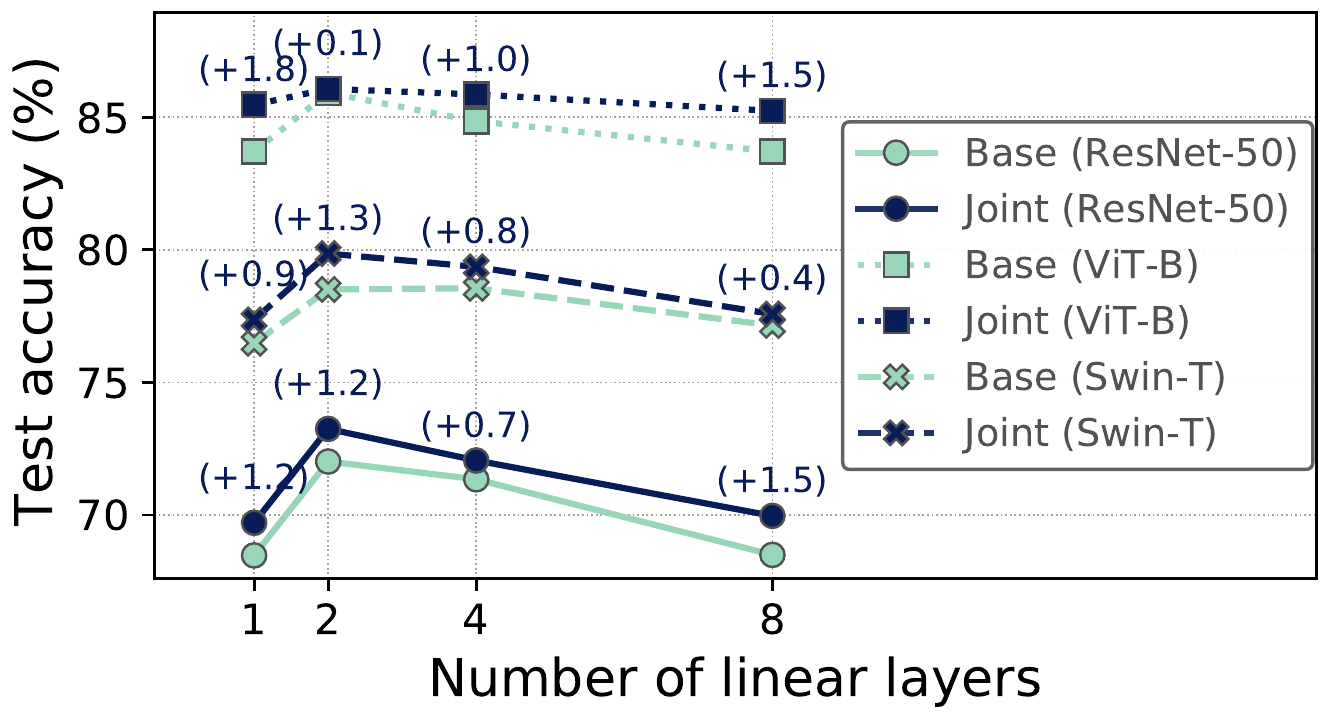}
  \caption{Effect of downstream MLP sizes.}
  \label{fig:ablation_mlpsize}
\end{subfigure}
\begin{subfigure}{.33\textwidth}
  \centering
  \includegraphics[scale=0.4]{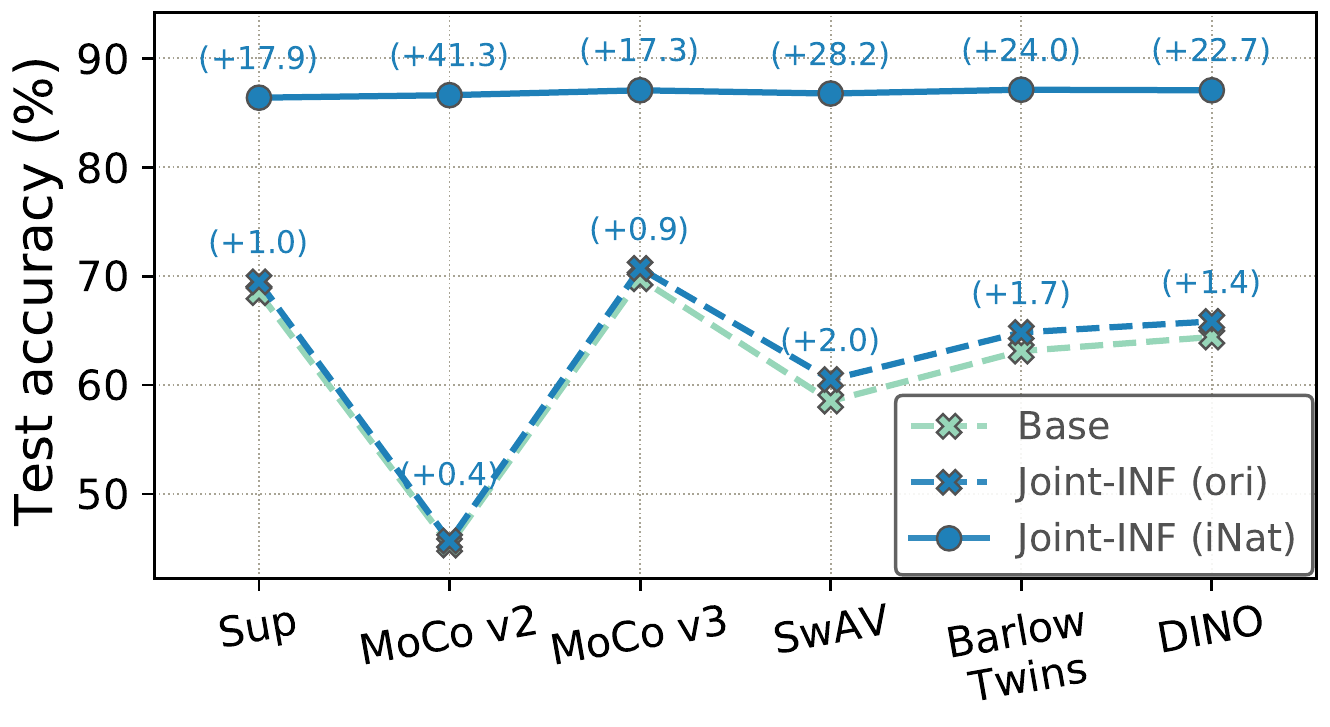}
  \caption{Effect of \knn{} features.}
  \label{fig:ablation_knnfeat}
\end{subfigure}
\vspace{-0.2cm}
\caption{Backbone architectures ablation results. We change the architectures of the (a) feature extractors, (b) \base{} classifier, and (c) \knn{} feature extractors. 
All results are for \cub{}~\test{} split with supervised pre-training objectives on \imagenet{}. Fig.~\ref{fig:ablation_knnfeat} uses \rn{}-50. 
($\cdot$) denotes the differences to \base{} under each variation.
}
\vspace{-0.4cm}
\label{fig:ablation_arch}
\end{figure*}

\paragraph{\joint{} is helpful in low-data regimes} Next, we analyze the impact of the proposed method when varying the size (by random sampling) of
(1) \base{} training data, (2) \knn{} datastore, (3) labeled pre-training dataset using supervised pre-training methods with both \imagenet{} and \inat{}.
\cref{fig:ablation_size} summarizes the results. 
Increasing the downstream data or \knn{} datastore sizes monotonically improves the performance of \joint{}. 
We also observe that with a better feature extractor (using \inat{} as pre-training dataset in this case), \joint{} is more robust to data size variation:
(1) using 60$\%$ of data for either \base{} (\cref{fig:ablation_size}a) or \knn{} (\cref{fig:ablation_size}b), \joint{} can achieve comparable results with the counterparts that use the full data, and
(2) with less than 20$\%$ of \inat{}, \joint{} could nearly match the \base{} results using the full pre-training data (\cref{fig:ablation_size}c).

\paragraph{\joint{} consistently outperforms \base{} over different pre-training backbone choices} 
\cref{fig:ablation_backbone} shows ablation results using different pre-training backbone architectures. \joint{} is better than \base{} across all backbone choices. We also note that the advantages of \joint{} do not diminish as the number of parameters of backbone gets larger for \rn{} based backbones, yet diminish for \swin{} (possibly due to overfitting on the downstream training set.)

\paragraph{\joint{} generalizes well to \base{} beyond linear classifier}
A natural question to ask is instead of the proposed method, what if we simply use a better classifier? We investigate this by increasing the single linear classifier to a multilayer perceptron (MLP) and varying the number of layers ($\{2, 4, 8\}$).
We report the results of \base{} and \joint{} of three different feature extractors in \cref{fig:ablation_mlpsize}.
We can see that the performance gains of \joint{} extend to multiple layers, even for MLP with 8 layers where the \base{} is overfitting on the training set.

\paragraph{Better \knn{} features result in larger performance gains}
We also ablate on the features used for \knn{} and report \jointinf{} results in \cref{fig:ablation_knnfeat}.
We choose the supervised pre-training method on \inat{} since it transfers better to \cub{} as shown in~\cref{fig:ablation_size} and achieves 86$\%$ top-1 accuracy with \knn{}.
The results show that further improvements may be possible by only swapping \knn{} features, suggesting that rather than re-train a new linear classifier with the improved feature representations, we can change the \knn{} datastore and save learning / training time.

\paragraph{Limitations}
In this paper, we use a vanilla implementation of \knn{}, which will incur large computational overhead when dealing with billions of training images.
One possible solution is to use implementations such as FAISS~\cite{johnson2019billion} for fast nearest neighbor search in high dimensional spaces.

\paragraph{The generality and flexibility of integrating \knn{}}
In this work, we propose to augment standard neural network classifiers with \knn{}.
Our results reveal a number of important benefits of incorporating \knn{}.
In terms of generality, we observe improved performance in every experimental setup: low \vs high data regimes used for pre-training, \base{}, or \knn{}; different architecture choices of \base{} (linear \vs~MLP); and classification tasks (binary \vs~multi-class, fine-grained \vs~standard \imagenet{}, species \vs other natural world tasks). 
In terms of flexibility, our hybrid approach is effective with diverse visual representations ranging from supervised to self-supervised methods, from \rn{} to \vit{} and \swin{}, from pre-trained on \imagenet{} to \inat{}.
We therefore hope our work could inspire and facilitate future research that reconsiders the role of the classical methods with the continuous advancement of the field.

\small{
\cvpara{Acknowledgement}
This work is supported by a Meta AI research grant awarded to Cornell University.
}

\appendix

\section{Supplementary Results and Discussion}
\label{supsec:result}

We provide the following items that shed further insight on the effect of \knn{} and the proposed \knn{} integration method:

\begin{itemize}[leftmargin=*]
    \item Further analysis of~\cref{sec:ana} from the main text (\cref{subsec:vis_supp});
    \item An extended discussion of the benefit of \knn{} and societal impact (\cref{subsec:dis_supp});
    \item Ablations experiments on the design choices of \knn{} (\cref{subsec:knn_ablation});
    \item Supplementary empirical results of \knn{} integration (\cref{subsec:results_supp,subsec:finetune_supp}).
\end{itemize}
Unless otherwise specified, all the experimental results of this section are on \cub{}~\test{} set, using the
supervised representations pre-trained on \imagenet{} with \rn{}-50 backbone.

\subsection{Further analysis of \knn{} integration}
\label{subsec:vis_supp}
\begin{figure}
\centering
\includegraphics[width=\columnwidth]{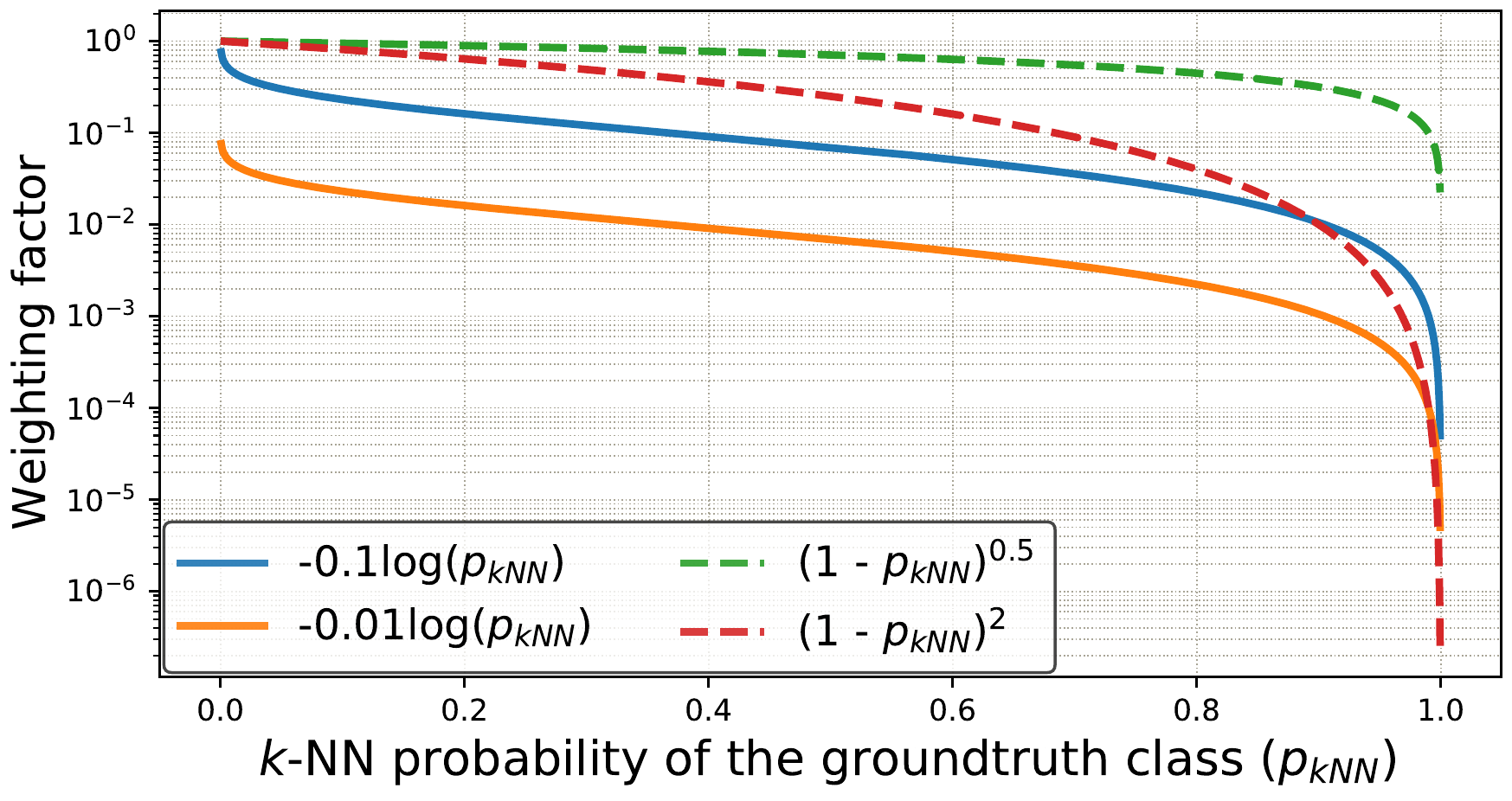}
\caption{
Visualizations of both negative log-likelihood (NLL) and focal loss (Focal) variants with different settings of $\alpha$ and $\gamma$.
Both variants reduce the relative loss for well-classified examples.
}
\label{fig:ana_weight}
\end{figure}

\begin{figure*}[t]
\centering
\includegraphics[width=\textwidth]{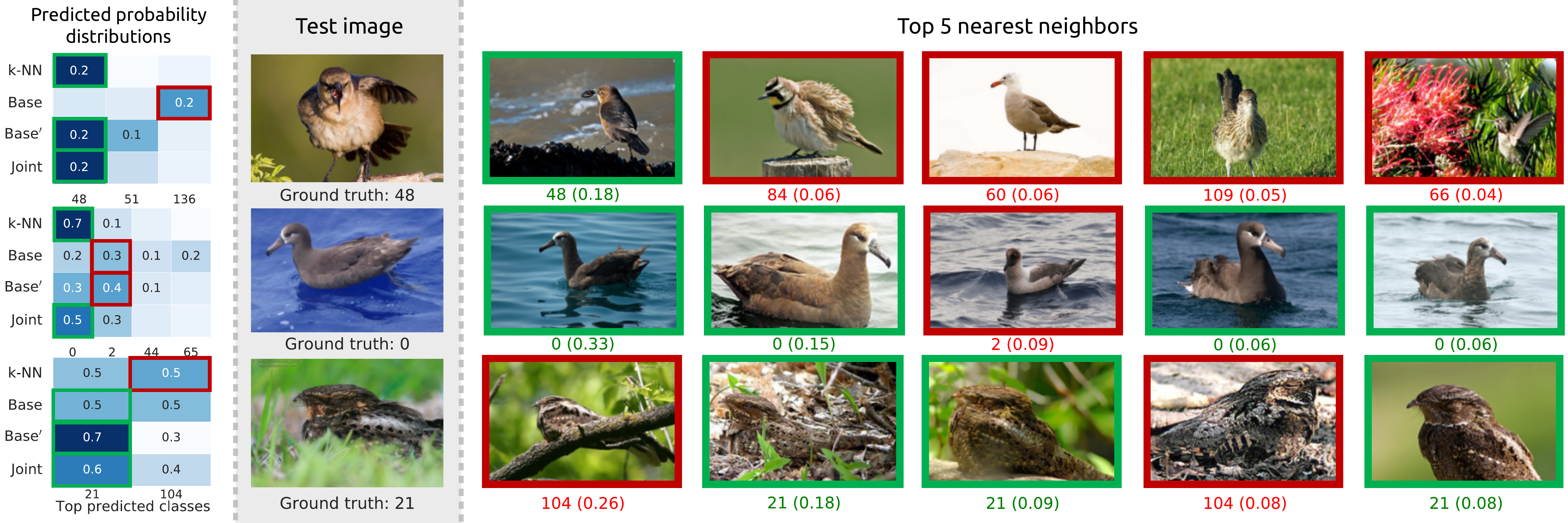}
\caption{
Additional results where \knn{} and \base{} produce different predictions. We show the predicted probability distribution (left), the test image (middle), and the top-5 nearest neighbors from the training set on the right. 
Only predicted probability $>0.1$ are presented in the left due to space constraint. 
Title of each retrieved samples follow the format: class index ($\exp{\left(-d(\cdot, \cdot) / \tau \right)}$ from~\cref{eq:knn_prob})
}
\label{fig:qual_supp}
\end{figure*}

\suppparagraph{Modulating factor initiations}
We visualize both modulating factor initiations (negative log-likelihood (NLL) and focal loss (Focal)) from~\cref{sec:ana} in~\cref{fig:ana_weight} with selected settings of $\alpha$ and $\gamma$.
NLL gives more weight for $p_{k\text{NN}} < 0.1$. 
Given the predicted distribution of \knn{} is sparser than \base{} (see~\cref{fig:qual,fig:qual_supp} for examples), NLL produces larger weights for the misclassified examples. This could be the reason why NLL variant performs slightly better than Focal in~\cref{tab:focal}.

\suppparagraph{More qualitative results}
\cref{fig:qual_supp} presents more \test{} examples of \cub{} where \knn{} and \base{} disagree with each other.
By explicitly memorizing the training data, 
\knn{} can produce correct predictions when \base{} produces low confidence false positives.
\cref{fig:qual_supp} also demonstrates the effect of proposed joint loss $\mathcal{L}_J$: the augmented linear classifier (\baseprime{}) creates higher confidence predictions compared to \base{}.

\subsection{Extended discussion}
\label{subsec:dis_supp}

\begin{figure*}
\centering
\includegraphics[width=\textwidth]{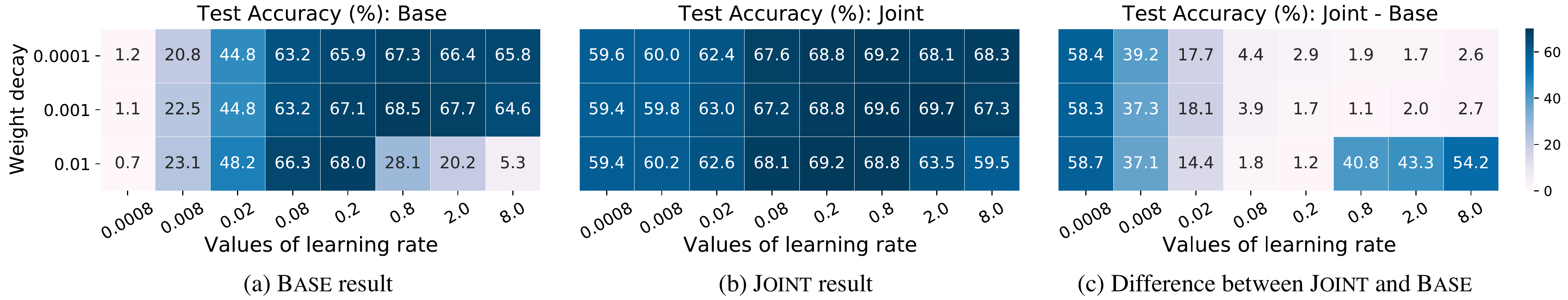}
\vspace{-0.5cm}
\caption{
\base{}, \joint{} and \joint{} - \base{} performance over different values of learning rate and weight decay using \cub{}.
\joint{} is robust to hyperparameter tuning.
}
\vspace{-0.2cm}
\label{fig:hp_supp}
\end{figure*}

\suppparagraph{\joint{} is robust to hyper-parameter tuning}
\cref{fig:hp_supp} presents performance of \base{} and \joint{} with different values of learning rate and weight decay.
As can be seen, \joint{} achieves relatively similar accuracy rates with a wide range of learning rates and weight decay values. 
This suggests that by incorporating \knn{}, the proposed method can be robust to the hyperparameters of \base{}, hence potentially reducing the range of grid search and increasing model development efficiency.

\begin{figure*}
\begin{subfigure}{.24\textwidth}
  \centering
  \includegraphics[scale=0.7]{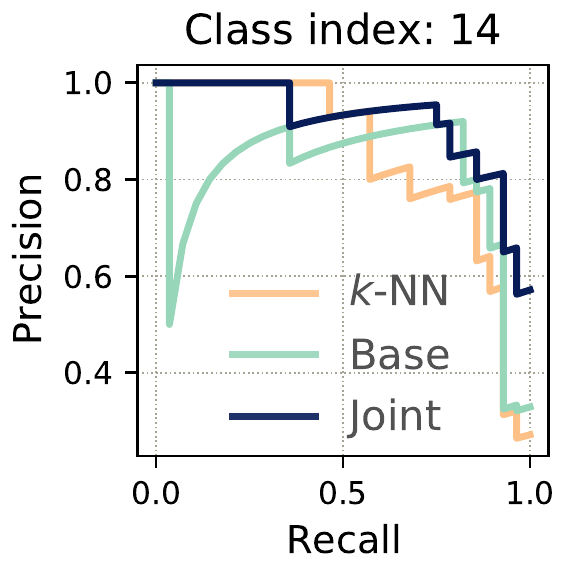}
  \caption{}
  \label{fig:pr_14}
\end{subfigure}%
\begin{subfigure}{.24\textwidth}
  \centering
  \includegraphics[scale=0.7]{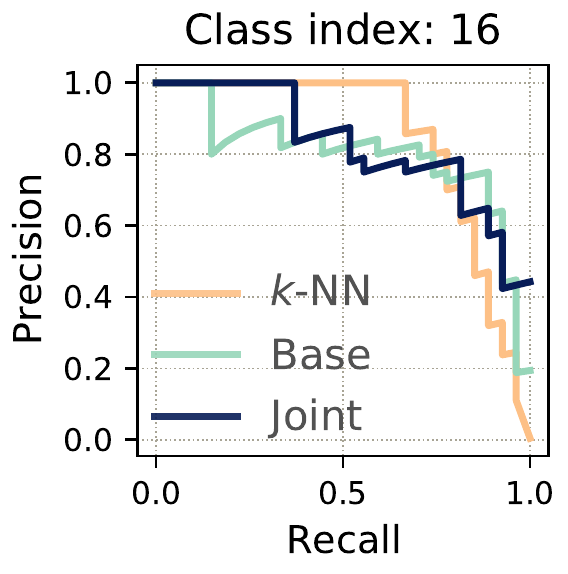}
  \caption{}
  \label{fig:pr_16}
\end{subfigure}
\begin{subfigure}{.24\textwidth}
  \centering
  \includegraphics[scale=0.7]{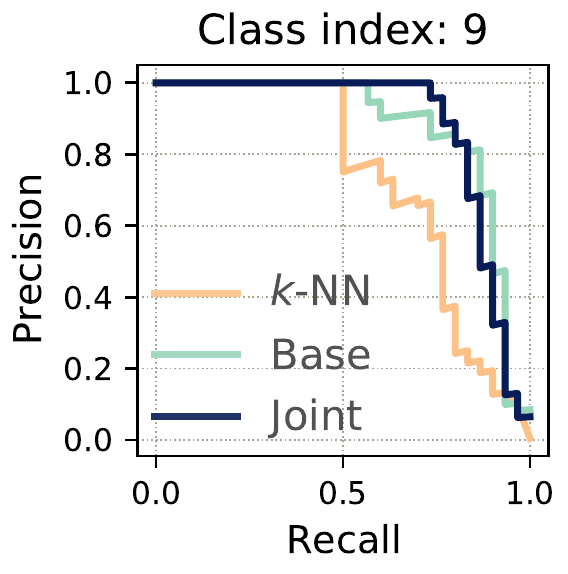}
  \caption{}
  \label{fig:pr_9}
\end{subfigure}
\begin{subfigure}{.24\textwidth}
  \centering
  \includegraphics[scale=0.7]{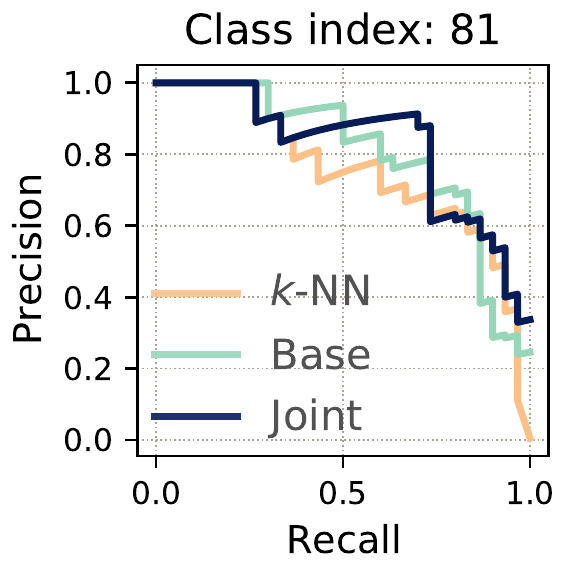}
  \caption{}
  \label{fig:pr_81}
\end{subfigure}
\caption{Precision-Recall (PR) curves of four classes in \cub{}. \base{} and \knn{} have complementary strength.
}
\label{suppfig:comp}
\end{figure*}

\suppparagraph{The complementary effect of \base{} and \knn{}}
As discussed in the main text, \knn{} has complementary strengths when compared to the neural network methods (lazy \vs~eager learners).
\Cref{sec:exp,sec:ana,sec:dis} present extensive empirical results to demonstrate that our proposed hybrid approach outperforms the \base{} and \knn{} counterparts.
We further examine the complementary strengths of both learning methods by plotting per-class Precision-Recall (PR) curves for \cub{} in~\cref{suppfig:comp}.
\cref{fig:pr_14,fig:pr_16} show \knn{} reaches higher recalls without any false positive predictions,
while \cref{fig:pr_9,fig:pr_81} present two cases where \base{} performs better.

\suppparagraph{Societal impact}
This paper investigates the role of ``old-school'' methods (\knn{}) in the era of the deep-learning revolution. 
Given a photo, the proposed method predicts probability distribution based on both learned global and local statistics in the training set, from the neural network and \knn{}, respectively.
Thus the learned statistics could reflect and amplify potential biases in the training data, such as \emph{selection, framing and labeling biases}~\cite{torralba2011unbiased}.
We would like to raise awareness of such issues. 
Any machine learning down-stream tasks should always apply fairness into consideration during algorithm development.
Note previous works have shown another benefit of \knn{}: improving interpretability~\cite{papernot2018deep,orhan2018simple,Junbo2019racnn}.
This suggests that integrating \knn{} could potentially help down-stream applications towards responsible and ethical machine learning practices.


\begin{figure*}
\centering
\includegraphics[width=0.95\textwidth]{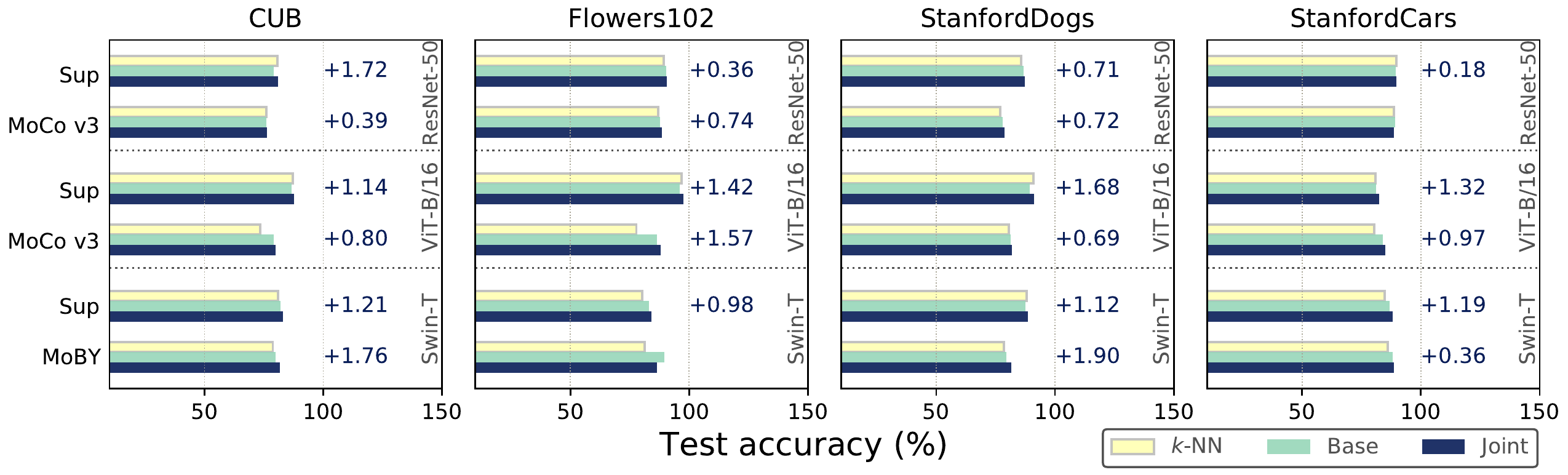}
\caption{
Fine-tuning results on the fine-grained down-stream datasets. All features are pre-trained on \imagenet{}. The y-axis denotes the pre-training objectives while the text on the right specify the feature extractor backbones. We also include the performance gains of \joint{} compare to \base{} for each representation evaluated if applicable.
}
\label{suppfig:fgvc_results}
\end{figure*}

\subsection{Other design choices of \knn{}}
\label{subsec:knn_ablation}

We present the preliminary experiments that lead to the design choices we made for the \knn{} classifier described in~\cref{subsec:knn}. 
We report the accuracy rate (\%) of \cub{} \test{} split.

\suppparagraph{Does the similarity metric of \knn{} matter?}
We use the negative squared Euclidean distance ($-d_{L2}$)\footnote{To avoid square root} to measure the similarity between the query $\vec{q}$ and examples in the memory bank $\mathcal{M}$. 
Another common metric is cosine similarity ($d_{cos}$), which is commonly used in the contrastive self-supervised learning literature (\cite{he2020moco, caron2021dino}, to name a few). We present \knn{} performance using both metrics with hyperparameters $k,\tau$ below.
\begin{center}
\small
\begin{tabular}{c  c c  }
Similarity metric & $-d_{L2}$ & $d_{cos}$ \\
\midrule
Accuracy (\%) &\textbf{63.67}  & 63.33\\
$k$  &128  &128 \\
$\tau$  &0.06 &0.1
\end{tabular}
\end{center}
Both similarity metrics lead to \emph{near the same} \knn{} classification results with different optimal $\tau$, possibly due to the fact that both metrics are related with the $l_2$ normalized features ($-d_{L2} = 2d_{cos} - 2$).
We thus adopt $-d_{L2}$ in the main text, expecting similar performance with $d_{cos}$.

\suppparagraph{\rn{}-based features: which down-sampling method to use?}
We experiment on different pooling methods for~\rn{}-based features: (1) \emph{MaxPool}, (2) \emph{AvgPool}, (3) \emph{R-MAC}~\cite{gordo2017end} (4) None: we simply flatten the feature map. 
The following are the \knn{} classification results, showing that \emph{MaxPool} outperforms the other pooling variants.
\begin{center}
\small
\begin{tabular}{l | c c c c }
Pooling method &\emph{MaxPool} &\emph{AvgPool} &\emph{R-MAC} &None \\
\hline
Accuracy (\%) &\textbf{63.67} &55.07 &60.56 &47.96 \\
\end{tabular}
\end{center}

\suppparagraph{\rn{}-based features: which layer's information to use as $\vec{x}$?}
We use the \emph{MaxPool}-ed feature maps from the last convolutional block of \rn{}-50 as input $\vec{x}$ for \knn{}.
The table below shows the \knn{} accuracy with feature maps from different convolutional blocks of \rn{}-50.
\begin{center}
\small
\begin{tabular}{c  c  c c c c}
Conv blocks &1 &2 &3 &4 &5 \\
\midrule
Accuracy (\%) &6.23 &9.51 &19.26 &35.47 &\textbf{63.67}
\end{tabular}
\end{center}
\knn{} performs the best when using the last feature map as input, as a deep neural network close to the output layer contain holistic semantic information of the image~\cite{orhan2018simple}.


\subsection{Additional results}
\label{subsec:results_supp}

\suppparagraph{Numerical results of \newt{} and fine-grained recognition experiments}
The \href{https://github.com/KMnP/nn-revisit}{project page} includes tables
presenting full empirical results including performance on the \val{} split and corresponding hyper-parameters used ($\tau, k, \alpha, \lambda$, learning rate, and weight decay).
These files can be read in conjunction with the figures and tables in the main text (\cref{fig:newt_results,fig:fgvc_results,tab:imgnet_results,fig:ablation_size,fig:ablation_arch}).
Note that we use in-house baselines instead of copying results from prior work for fair-comparison purposes.
All the experiments are trained using the same grid search range, validation set, learning rate schedule, input augmentation schemes \etc.
See~\cref{supsec:detail} for details.

\suppparagraph{\imagenet{} results with the off-the-shelf \base{}}
We report results of in-house \base{} for fair comparison across all feature configurations in~\cref{subsec:imgnet_exp}.
For completeness, we additionally compare the \knn{} integration (during test time only) with off-the-shelf \imagenet{} classifiers.
\cref{supptab:imgnet_results} summarizes the results.
As can be seen, incorporating \knn{} in test time outperforms the vanilla neural network based classifier counterpart across all the evaluated features.

\begin{table}
\scriptsize
\begin{center}
\resizebox{\columnwidth}{!}{%
\begin{tabular}{l l  l l l }
\toprule
\multirow{2}{*}{\textbf{Backbone}} 
&\multirow{2}{*}{\shortstack[l]{\textbf{Pre-trained}\\\textbf{Objective}}} 
&\multicolumn{3}{ c }{\textbf{Top 1 Accuracy ($\%$)}} \\
\cline{3-5}\noalign{\smallskip}
& & \textbf{\knn{}}  &\textbf{\base{}} &\textbf{\jointinf{}} 
\\
\midrule
\multirow{4}{*}{\rn{}-50}
&\suplong{} &72.65 &76.01 &\Rise{0.37}\textbf{76.39} \\

&\moco{} (ep100)  &61.97  &68.96 &\Rise{1.06}\textbf{70.02}  \\

&\moco{} (ep300)  &66.51  &72.87 &\Rise{0.99}\textbf{73.86} \\

&\moco{} (ep1000) &67.33  &74.57	&\Rise{0.96}\textbf{75.53}  \\
\midrule

\vit{}-B/16 &\moco{}  &71.22	&76.61	&\Rise{0.91}\textbf{77.52} \\

\vit{}-S/16 &\moco{}  &67.55	&73.33	&\Rise{1.15}\textbf{74.48} \\

\bottomrule
\end{tabular}
}
\caption{Results on \imagenet{} using the off-the-shelf weights.}
\vspace{-0.5cm}
\label{supptab:imgnet_results}
\end{center}
\end{table}

\subsection{End-to-end fine-tuning}
\label{subsec:finetune_supp}

\suppparagraph{Settings}
We additionally run experiments under fine-tuning transfer learning protocol, where we use the pre-trained models as initiations for the four fine-grained object classification tasks~\cite{nilsback2008automated,WahCUB_200_2011,parkhi2012dogs,gebru2017cars}.
The parameters of the feature extractors $\phi(\cdot)$ are modified during training.
We thus update the memory bank $\mathcal{M}$ of \knn{} every epoch for \joint{} method.
Similar to experiments in~\cref{subsec:fgvc_exp}, we use either publicly available or randomly sampled \val{} set to select hyperparameters.
All feature representations are pre-trained on \imagenet{}.

\suppparagraph{Results}
\cref{suppfig:fgvc_results} presents the \test{} set accuracy rates of \knn{}, \base{}, \joint{}.
We observe that \joint{} could improve the vanilla neural network model.
For example, \joint{} achieves 1.72 $(2.17\%)$ accuracy gain using imagenet supervised representations with \rn{}-50 backbone.

\begin{figure}[t]
\centering
\includegraphics[width=0.95\columnwidth]{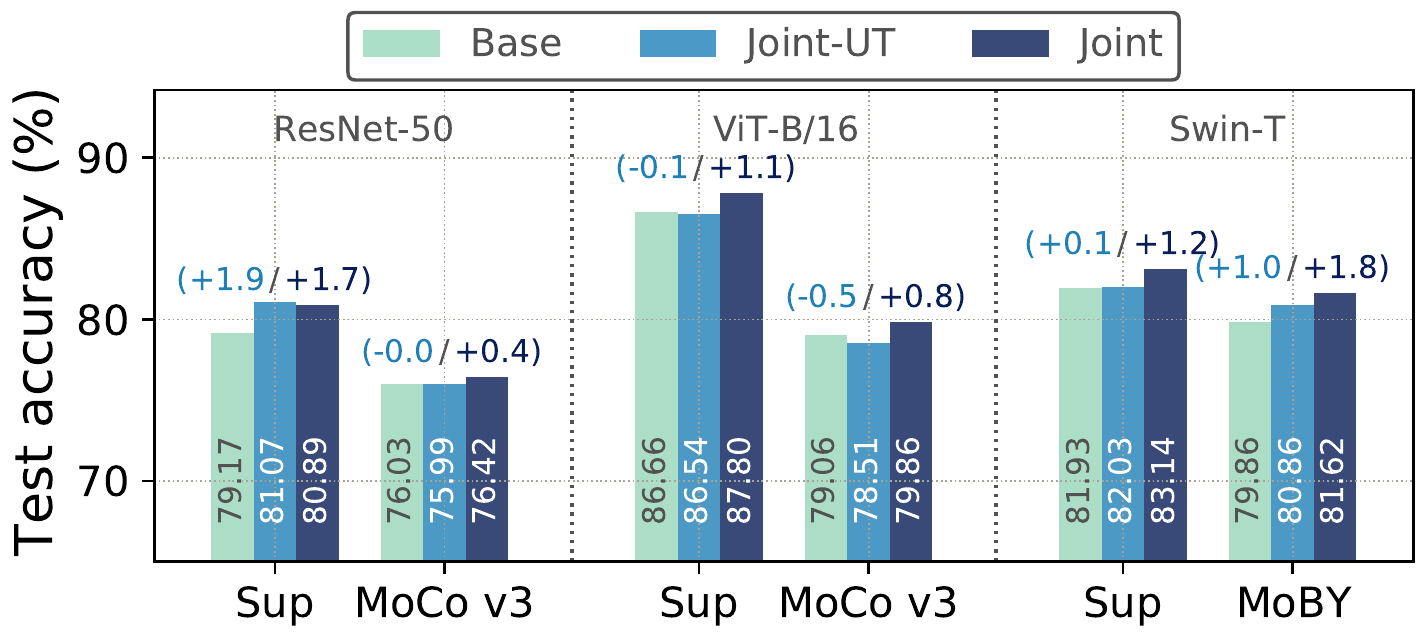}
\caption{
Effect of updating memory bank. 
The x-axis denotes the pre-training objectives while the text on the top specify the feature extractor backbones.
We also include the difference of \joint{}\textsc{-ut} and \joint{} compared to \base{} for each representation evaluated.
}
\label{suppfig:ft_update}
\end{figure}

\suppparagraph{Updating $\mathcal{M}$} Under the fine-tuning protocol, we update the memory bank $\mathcal{M}$ every epoch to reflect the change of the feature extractor. 
We additionally experiment on a variant \joint{}\textsc{-ut}: \textbf{U}pdating $\mathcal{M}$ during \textbf{T}est time only. 
In particular, we use the pre-trained features for $\mathcal{M}$ during the end-to-end training, and the updated features during test time.
\cref{suppfig:ft_update} presents the results on \cub{}. 
Updating $\mathcal{M}$ every epoch (\joint{}) generalizes better than \joint{}\textsc{-ut}. 
Considering its slight computational advantage, \joint{}\textsc{-ut} may be preferred in cases where it is desirable to trade off a certain amount of accuracy for the reduced training time.
\eg, for the \cub{} dataset, \joint{}\textsc{-ut} takes around 6 hours (20.8\%) less than \joint{} with the \swin{}-T supervised \imagenet{} pre-training features when trained on a single A100 GPU using our PyTorch implementation.

\section{Reproducibility Details}
\label{supsec:detail}

\subsection{Implementation details}
\label{supsec:imp}

We use Pytorch~\cite{paszke2017pytorch} 
to implement and train all the models on NVIDIA Tesla V100 and A100-40GB GPUs, for pre-training and transfer learning experiments, respectively.
We adopt standard image augmentation strategy during training (randomly resize crop to 224 $\times$ 224 and random horizontal flip). Other implementation details are described below and the specific hyper-parameter values for each experiment can be found in the project page mentioned in~\cref{subsec:results_supp}.

\suppparagraph{Modulating factor of $\mathcal{L}_J$}
Similar to pytorch's BCELoss\footnote{\href{https://pytorch.org/docs/stable/generated/torch.nn.BCELoss.html}{pytorch documentation}}, we clamp the $log$ function outputs to be greater than or equal to -100, so we can always have a finite value of the modulating factor.
\cref{supptab:joint_imp} presents the range of other hyper-parameters used for \knn{} and \joint{}.

\suppparagraph{Transfer learning settings} 
Since the dataset is not balanced, we follow~\cite{focal_loss} to stabilize the training processing by initializing the the bias for the last linear classification layer  with $b = -\log\left(\left(1 - \pi\right)/\pi\right)$, where the prior probability $\pi$ is set to 0.01.
\cref{supptab:linear_imp,supptab:ft_imp} summarize the optimization configurations we used.
Following~\cite{wslimageseccv2018}, we conduct a coarse grid search to find the learning rate and weight decay values using \val{} split.
The learning rate is set as $\text{Base LR} / 256 \times b$, where $b$ is the batch size used for the particular model, and Base LR is chosen from the range specified in~\cref{supptab:linear_imp,supptab:ft_imp}.

\suppparagraph{Pre-training settings}
We additionally trained three supervised \rn{}-50 models using subsets of \imagenet{}.
These models are trained on 8 GPUs.
We use a batch size of 1024 images,
stochastic gradient descent with a momentum of 0.9, and a weight decay of 0.0001.
The base learning rate is set according to $0.1 / 256 \times b$, where $b$ is the batch size used for the particular model. 
The learning rate is warmed up linearly from 0 to the base learning rate during the first $5\%$ of the whole iterations, then reduced by a factor of $0.1$ during epoch $\{30, 60, 90\}$.

\subsection{Datasets and feature representations}
\label{suppsec:tasks}

The statistics of the evaluated tasks and the associated datasets are shown in~\cref{tab:datasets}.
Features of \newt{} are obtained from the researchers who propose the datasets.
\cref{tab:features} enumerates all the features we used in the main text.

\begin{table}
\small
\begin{center}
\begin{tabular}{l l}
\toprule
\textbf{Config}
&\textbf{Values}
\\
\midrule
$k$ &$\{2^{\{0:1:9\}}, mean\}$
\\
$\tau$ &$\{$0.001, 0.01, 0.1, 1, 10, \\
&$0.01 \times \{1:1:10\}, 0.1 \times \{1:1:10\}$
$\}$
\\
$\alpha$ &$\{0.01, 0.001, 0.0001\}$
\\
$\lambda$  &$\{0.50:.05:0.95\}$
\\
\bottomrule
\end{tabular}
\caption{\joint{} setting. $mean$ represents the average image count per class for different tasks.}
\label{supptab:joint_imp}
\end{center}
\end{table}

\begin{table}
\small
\begin{center}
\begin{tabular}{l l}
\toprule
\textbf{Config}
&\textbf{Value}
\\
\midrule
optimizer &SGD
\\
base LR range &$\{$0.5, 0.25, 0.1, 0.05, 0.025, \\
&0.01, 0.0025, 0.001
$\}^{\star}$
\\
weight decay range &$\{0.01, 0.001, 0.0001, 0.0\}$
\\
optimizer momentum  &0.9
\\
learning rate schedule &cosine decay
\\
warm up epochs &10
\\
total epochs &100
\\
\bottomrule
\end{tabular}
\caption{Linear evaluation setting. 
$^{\star}$For the self-supervised features, we use a different range for base learning rate: $\{40, 20, 10, 5.0, 2.5, 1.0, 0.5, 0.05, 0.1\}$ since the feature magnitudes are different from the supervised variants. Note that we can alternatively add a normalization layer before the linear classifier following~\cite{he2021mae} instead of using larger learning rate values. 
}
\label{supptab:linear_imp}
\end{center}
\end{table}

\begin{table}
\small
\begin{center}
\begin{tabular}{l l}
\toprule
\textbf{Config}
&\textbf{Value}
\\
\midrule
optimizer &SGD
\\
base LR range &$\{$0.05, 0.025, 0.005, 
\\
&0.0025, 0.0005, 0.00025
$\}$
\\
weight decay range &$\{0.01, 0.001, 0.0001, 0.00001\}$
\\
optimizer momentum  &0.9
\\
learning rate schedule &cosine decay
\\
warm up epochs &5
\\
total epochs &300
\\
\bottomrule
\end{tabular}
\caption{End-to-end fine-tuning evaluation setting.}
\label{supptab:ft_imp}
\end{center}
\end{table}

\begin{table*}[h]
\small
\begin{center}
\resizebox{\textwidth}{!}{%
\begin{tabular}{l l  l l l l l}
\toprule
\textbf{Dataset}   &\textbf{Description}  & \textbf{\# Classes}    &\textbf{Train}  &\textbf{Val}  &\textbf{Test} &\textbf{License with links}\\ 
\midrule
\imagenet{}~\cite{imagenet_cvpr09}  &Image recognition
&1,000 & 1,281,167 &50,000 & - &\href{https://www.image-net.org/download}{Non-commercial} \\
\midrule

\longcub{}~\cite{WahCUB_200_2011}
& Fine-grained bird species recognition
&200
&5,394$^{\star}$	&600$^{\star}$ &5,794	
&\href{http://www.vision.caltech.edu/visipedia/CUB-200-2011.html}{Non-commercial} \\

\flowers{}~\cite{nilsback2008automated}
& Fine-grained flower species recognition
&102
&1,020	&1,020	&6,149 
&\href{https://www.robots.ox.ac.uk/~vgg/data/flowers/}{Not specified}\\

\dogs{}~\cite{parkhi2012dogs}
 &Fine-grained dog species recognition  &120 
 &10,800$^{\star}$	&1,200$^{\star}$	&8,580 
 &\href{http://vision.stanford.edu/aditya86/ImageNetDogs/}{Not specified}\\

\cars{}~\cite{gebru2017cars}
& Fine-grained car detection  &196  
&7,329$^{\star}$	&815$^{\star}$	&8,041 
&\href{http://ai.stanford.edu/~jkrause/cars/car_dataset.html}{Non-commercial}
\\

\midrule

Natural World Tasks (\newt{})~\cite{VanHorn2021newt}
&A set of 196 natural world classification tasks
& & & & 
&\href{https://github.com/visipedia/inat_comp/tree/master/2021}{Non-commercial}\\
\quad Appearance-Age &\quad 14 binary tasks &2	&\meanstd{114.29}{34.99}	&- &\meanstd{114.29}{34.99} \\
\quad Appearance-Attribute &\quad 7 binary tasks &2	&\meanstd{97.71}{5.60}	&- &\meanstd{87.43}{24.85} \\
\quad Appearance-Health &\quad 9 binary tasks	&2 &\meanstd{122.22}{41.57} &-	&\meanstd{116.44}{45.15} \\
\quad Appearance-Species &\quad 102 binary tasks	&2 &\meanstd{97.29}{25.93} &-	&\meanstd{100.12}{25.19}  \\
\quad Behavior &\quad 16 binary tasks &2	&\meanstd{100.00}{0.00}	&- &\meanstd{88.06}{18.10} \\
\quad Context &\quad 8 binary tasks &2 &\meanstd{100.00}{0.00}	&- &\meanstd{105.50}{25.82} \\
\quad Counting &\quad 2 binary tasks &2 &\meanstd{100.00}{0.00} &-	&\meanstd{100.00}{0.00} \\
\quad Gestalt &\quad 6 binary tasks &2 &\meanstd{350.00}{111.80} &-	&\meanstd{349.83}{111.73}
\\

\Xhline{0.7pt}\noalign{\smallskip}
\end{tabular}
}
\caption{Specifications of the various datasets evaluated. Image number with $^{\star}$ are the subset we randomly sampled since no publicly data splits are available. 
We report mean and standard deviation of data sizes for \newt{}.
``Non-commercial'' is short for ``non-commercial research and educational purposes only''. 
}
\label{tab:datasets}
\end{center}
\end{table*}

\begin{table*}
\begin{center}
\resizebox{\textwidth}{!}{%
\begin{tabular}{l l l l l l l l}
\toprule
\textbf{\shortstack[l]{Pre-trained\\Backbone}}
&\textbf{Pre-trained Objective}
&\textbf{Pre-trained Data}
&\textbf{Throughput (img/s)}
&\textbf{\shortstack[l]{\# params\\(M)}} 
&\textbf{Feature dim} &\textbf{Batch Size}
&\textbf{\shortstack[l]{Pre-trained\\Model}}
\\  
\midrule
\multirow{6}{*}{\rn-50~\cite{he2016rn} }
&\multirow{9}{*}{\suplong{} }
&\imagenet{}~\cite{imagenet_cvpr09}

&\multirow{6}{*}{1581.6}
&\multirow{6}{*}{23}
&\multirow{6}{*}{2048}
&\multirow{6}{*}{2048 / 384}
&\href{https://pytorch.org/vision/stable/models.html}{link}
\\

&
&\imagenet{}-25$\%$~\cite{imagenet_cvpr09}
&&&&
&in house
\\
&
&\imagenet{}-50$\%$~\cite{imagenet_cvpr09}
&&&&&in house
\\
&
&\imagenet{}-75$\%$~\cite{imagenet_cvpr09}
&&&&&in house
\\
&
&iNaturalist2021 (iNat)~\cite{VanHorn2021newt}
&&&&
&\href{https://cornell.app.box.com/s/bnyhq5lwobu6fgjrub44zle0pyjijbmw}{link}
\\

&
&iNaturalist2021-mini (iNat-mini)~\cite{VanHorn2021newt}
&&&&
&\href{https://cornell.app.box.com/s/bnyhq5lwobu6fgjrub44zle0pyjijbmw}{link}
\\

\cmidrule{3-8}
\rn{}-18~\cite{he2016rn}
&&\imagenet{}~\cite{imagenet_cvpr09}
&4602.5
&11 &\multirow{3}{*}{2048}
&\multirow{3}{*}{2048 / -}
&\href{https://pytorch.org/vision/stable/models.html}{link}
\\
\rn{}-101~\cite{he2016rn}
&&\imagenet{}~\cite{imagenet_cvpr09}
&963.6
&43 &&
&\href{https://pytorch.org/vision/stable/models.html}{link}
\\
\rn{}-152~\cite{he2016rn}
&&\imagenet{}~\cite{imagenet_cvpr09}
&685.0
&58 &&
&\href{https://pytorch.org/vision/stable/models.html}{link}
\\

\cmidrule{2-8}
\multirow{5}{*}{\rn-50~\cite{he2016rn} }
&\mocotwo{}~\cite{chen2020mocov2}  
&\multirow{5}{*}{\imagenet{}~\cite{imagenet_cvpr09}}
&\multirow{5}{*}{1581.6}
&\multirow{5}{*}{23}
&\multirow{5}{*}{2048}
&\multirow{5}{*}{2048 / 384}
&\href{https://github.com/facebookresearch/moco}{link}
\\

&\moco{}~\cite{chen2021mocov3} 
&&&&&&\href{https://github.com/facebookresearch/moco-v3}{link}
\\
&\swav{}~\cite{caron2020swav}  
&&&&&&\href{https://github.com/facebookresearch/swav}{link}
\\
&\dino{}~\cite{caron2021dino}  
&&&&&&\href{https://github.com/facebookresearch/dino}{link}
\\
&\btwins{}~\cite{zbontar2021barlow}
&&&&&&\href{https://github.com/facebookresearch/barlowtwins}{link}
\\


\midrule
\vit-B/16~\cite{dosovitskiy2020vit}

&\multirow{2}{*}{\suplong{} }
&\multirow{2}{*}{\imagenet{}~\cite{imagenet_cvpr09}}
&1264.7 & 85 & 768 &1024 / 192
&\href{https://console.cloud.google.com/storage/browser/vit_models;tab=objects}{link}
\\

\vit-L/16~\cite{dosovitskiy2020vit}
&&
&460.2 &307 & 1024 &512  / -
&\href{https://console.cloud.google.com/storage/browser/vit_models;tab=objects}{link}
\\
\cmidrule{2-8}

\vit-S/16~\cite{dosovitskiy2020vit}
&\dino{}~\cite{caron2021dino}  
&\multirow{4}{*}{\imagenet{}~\cite{imagenet_cvpr09}}
&2342.0 & 21 & 384 &1024 / -
&\href{https://github.com/facebookresearch/dino}{link}
\\
\vit-S/16~\cite{dosovitskiy2020vit}
&\moco{}~\cite{chen2021mocov3} &
&2342.0 & 21 & 384 &1024 / -
&\href{https://github.com/facebookresearch/moco-v3}{link}
\\
\vit-B/16~\cite{dosovitskiy2020vit}
&\dino{}~\cite{caron2021dino}  &
&1264.7 & 85 & 768 &1024 / -
&\href{https://github.com/facebookresearch/dino}{link}
\\
\vit-B/16~\cite{dosovitskiy2020vit}
&\moco{}~\cite{chen2021mocov3} &
&1264.7 & 85 & 768 &1024 / 192
&\href{https://github.com/facebookresearch/moco-v3}{link}
\\
\midrule

\swin-T~\cite{liu2021swin}
&\multirow{3}{*}{\suplong{} } 
&\multirow{3}{*}{\imagenet{}~\cite{imagenet_cvpr09}}
&1438.3 &29 &768 &1024 / 192
&\multirow{3}{*}{\href{https://github.com/microsoft/Swin-Transformer}{link}}
\\
\swin-S~\cite{liu2021swin}
&& &946.8	&50	&768 &1024 / -
&
\\
\swin-B~\cite{liu2021swin}
&& &716.9	&88	&1024 &1024 / -
&\\
\cmidrule{2-8}

\swin-T~\cite{liu2021swin} &\moby{}~\cite{xie2021moby}&\imagenet{}~\cite{imagenet_cvpr09}
&1438.2 &29 &768 &1024 / 192
&\href{https://github.com/SwinTransformer/Transformer-SSL}{link}
\\
\bottomrule
\end{tabular}
} 
\caption{Specifications of different representation learning approaches ($\phi(\cdot)$) evaluated for fine-grained object recognition and \imagenet{} classification tasks.
All features are pre-trained with an randomly resized cropped input image of size 224$\times$224.
The throughput (im/s) is calculated on a NVIDIA A100 GPU. 
Parameters (M) are of the feature extractor.
Batch size column reports the batch size for linear evaluation and fine-tuning experiments, respectively.
See~\cref{supsec:imp} and attached CSV files for more implementation details. }
\label{tab:features}
\end{center}
\end{table*}

\clearpage
\newpage
{\small
\bibliographystyle{ieee_fullname}
\bibliography{egbib}
}

\end{document}